\documentclass{article}
\usepackage{iclr2027_conference,times}
\iclrfinalcopy

\usepackage{amsmath,amsfonts,bm}

\def\eqref#1{equation~\ref{#1}}

\def\1{\bm{1}}

\DeclareMathAlphabet{\mathsfit}{\encodingdefault}{\sfdefault}{m}{sl}
\SetMathAlphabet{\mathsfit}{bold}{\encodingdefault}{\sfdefault}{bx}{n}

\usepackage{comment}

\usepackage{hyperref}
\usepackage{url}
\usepackage{graphicx}
\usepackage{wrapfig}
\usepackage{needspace}
\usepackage{booktabs}
\usepackage{tabularx}
\usepackage{float}
\usepackage{amsmath,amssymb}
\usepackage{multirow}
\usepackage{subcaption}
\usepackage{colortbl}
\definecolor{revrow}{rgb}{1.0,0.97,0.80}
\newcommand{\rev}[1]{#1}
\newcommand{\ns}[1]{\textcolor{gray}{#1}}
\newcommand{\ci}[2]{{\scriptsize\textcolor{gray}{[$#1$,\,$#2$]}}}

\newcommand{\tablelayout}{\footnotesize\setlength{\tabcolsep}{3pt}\renewcommand{\arraystretch}{1.05}}

\title{When the Score Becomes the Target: Rethinking Metric Validity in Autonomous Driving}

\author{%
\hypersetup{hidelinks}%
Morui Zhu\textsuperscript{1,2},
Deyuan Qu\textsuperscript{2},
Qi Chen\textsuperscript{2},
Kentaro Oguchi\textsuperscript{2},
Qing Yang\textsuperscript{1}\\[4pt]
\normalfont\textsuperscript{1} University of North Texas\\
\normalfont\textsuperscript{2} Toyota InfoTech Labs}

\definecolor{tableimprove}{rgb}{0.12,0.43,0.27}
\definecolor{tabledegrade}{rgb}{0.72,0.20,0.18}
\newcommand{\metrichead}[2]{\mbox{#1\hspace{2pt}$#2$}}

\begin{document}
\maketitle
\fancyhead{} 
\renewcommand{\headrulewidth}{0pt}

\begin{abstract}
Driving benchmark scores are increasingly used not only for evaluation but also as optimization targets.
This raises a fundamental question: do score gains remain reliable evidence of driving improvement once the score itself is optimized?
We address this question by examining how the scoring process responds to changes in driving behavior and whether the resulting gains persist under repeated execution and replanning.
We decompose the process into execution, measurement, subscore mapping, and aggregation.
Controlled interventions reveal substantial behavioral changes that receive little score response because distinctions are omitted, thresholded, or attenuated between requested and executed motion.
Closed-loop comparisons further show that optimization gains can reverse when the execution interface changes, demonstrating their dependence on how requests are executed and returned as feedback.
Together, these findings connect the behavioral distinctions preserved by a metric to the conditions under which its gains transfer.
Metric validity under optimization therefore requires examining both what the scoring process measures and how the optimized behavior is executed.

\end{abstract}

\section{Introduction}
\label{sec:intro}
\label{sec:pitfalls}

{\centering\itshape
\setlength{\parskip}{0pt}
Autonomous driving aims to transport people safely, comfortably, and efficiently to their destinations, while following traffic rules and interacting safely with other road users.
\par}

\begin{figure}[!t]
\vspace{-6pt}
\centering
\captionsetup{skip=6pt}
\includegraphics[width=\linewidth]{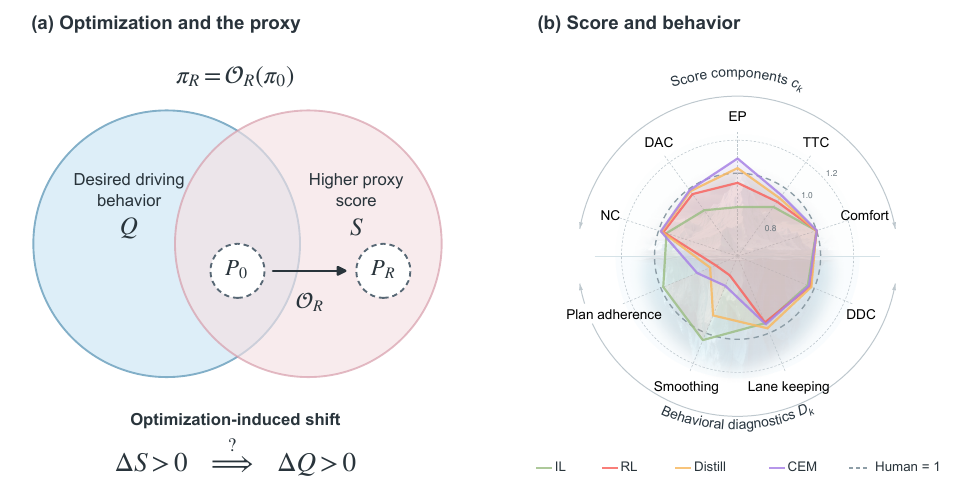}
\caption{\textbf{Optimization and score--behavior mismatch.} (a) Conceptual distributions: optimizing $R$ changes $P_0$ to $P_R$; higher $S$ does not imply better $Q$. (b) Group-mean score components $c_k$ and diagnostics $D_k$ for PDMS (Predictive Driver Model Score). Each axis is oriented so that a larger radius represents better compliance or a smaller behavioral error. Values are normalized relative to the logged-human mean, shown at radius 1, to place heterogeneous quantities on a common visual scale. The radial scale is nonlinear.}
\label{fig:teaser}
\vspace{-10pt}
\end{figure}

These goals motivate driving benchmarks \citep{caesar2021nuplan,dauner2024navsim,jia2024bench2drive,zhou2024hugsim} and learned planning systems \citep{liao2025diffusiondrive,zheng2025diffusionplanner,li2026recogdrive}.
Benchmark scores provide a common basis for comparing progress toward these goals and increasingly serve as rewards or trajectory-selection objectives \citep{nord2026,li2025gtrs,xu2026toad}.
Once a score becomes an optimization target, a central question is whether its gains continue to reflect improvements in the driving behaviors it is intended to measure, a property we refer to as \textit{metric validity}. 
Optimization changes the trajectories a planner produces, so a score's ability to distinguish behavior on existing outputs does not establish its validity on optimized outputs (Figure~\ref{fig:teaser}).
This concern is related to Goodhart's law~\citep{manheim2018categorizing,karwowski2024goodhart},
while prior driving studies have exposed evaluation shortcuts and discrepancies between open- and closed-loop performance~ \citep{zhai2023rethinking,dauner2023parting,wang2026openloop}.
Recent post-training results further show that open-loop gains do not always translate into closed-loop improvements~\citep{worldengine}.
These findings establish that score--behavior discrepancies can occur, but leave open where they arise in the driving evaluation process and how optimization exposes them.
We address this gap by examining behavioral distinctions obscured by score aggregation, thresholding, and transformations between planned and executed motion, and by tracing whether optimization gains persist under repeated replanning.
Our analysis and large-scale study provide practical guidance on which behaviors a metric should distinguish and how its validity should be reassessed after optimization.

\section{Score--Behavior Analysis}
\label{sec:validation}
\label{sec:measurement_target}

Let $S$ denote the reported benchmark score and $D_k$ a behavior-specific diagnostic relevant to an improvement claim.
Let $R$ denote the optimization objective, which may equal $S$, approximate it, or include auxiliary terms.
An optimization procedure $\mathcal{O}_R$ transforms a planner $\pi_0$ into $\pi_R=\mathcal{O}_R(\pi_0)$ through training or trajectory selection, inducing output distributions $P_{0,e}$ and $P_{R,e}$ under evaluation setting $e$ on a shared initial-scene distribution.
For any measured quantity $F$, we define
\begin{equation}
\Delta F_e
=
\mathbb{E}_{P_{R,e}}[F_e]
-
\mathbb{E}_{P_{0,e}}[F_e].
\label{eq:paired-change}
\end{equation}
Our central question is whether a score gain, $\Delta S_e>0$, is accompanied by improvement in the driving behaviors the score is intended to measure.
We assess this correspondence using diagnostics $D_k$ specific to the behavior of interest, rather than reducing driving quality to a single surrogate.

\textit{Metric validity} therefore depends on both the outputs and how they are evaluated: validity on $P_{0,e}$ does not by itself establish validity on $P_{R,e}$ or under a different execution or feedback setting.
For a request $u$ at a fixed initial state, with execution $T_e$, measurements $m_e$, subscore mappings $h_e$, and aggregation $G_e$, the score is
\begin{equation}
\bar S_e(u)
=
G_e\!\left(h_e\!\left(m_e\!\left(T_e(u)\right)\right)\right).
\label{eq:measurement-chain}
\end{equation}
This decomposition identifies three possible blind spots: omission, insensitive subscore mappings, and transformations between requested and scored motion.
For each, Section~\ref{sec:stress-testing} tests the predicted insensitivity and checks whether the corresponding behavior deteriorates in optimized outputs despite a score gain.

\paragraph{Aggregation and omission.}
\label{sec:omission}
Suppressing $e$, represent a measured behavior by a candidate component $c_k=h_k(D_k)$ and write $S=G(c_1,\ldots,c_K)$. On differentiable branches, holding other inputs fixed,
\begin{equation}
\frac{\partial S}{\partial D_k}
=\underbrace{\frac{\partial G}{\partial c_k}}_{\text{aggregation sensitivity}}
\underbrace{h_k'(D_k)}_{\text{subscore sensitivity}}.
\label{eq:sensitivity}
\end{equation}
Omission makes the first factor zero throughout the relevant range: changing this component alone cannot change $S$. Matched trajectory probes test this invariance while controlling other scored properties (Section~\ref{sec:blindspots}).
Even for included components, an aggregate gain can reflect trade-offs or changes in gate pass rates; component and conditional measurements distinguish these contributions (Section~\ref{sec:aggregation-evidence}).

\paragraph{Thresholding and saturation.}
\label{sec:thresholding}
An included component can instead have zero subscore sensitivity. For an upper-bounded quantity $d$,
\begin{equation}
c(d)=\mathbf1[d<b],\qquad c'(d)=0\quad\text{for }d<b.
\label{eq:flat-region}
\end{equation}
Values inside the passing region have equal scores when other inputs are fixed; clipping has analogous flat regions. Continuous readouts reveal changes within them, while rescoring fixed outputs with a different $h_e$ tests score resolution (Section~\ref{sec:nuplan}); finite differences handle boundaries.

\paragraph{Execution transformation.}
\label{sec:transformation}
Let $s_e=G_e\circ h_e\circ m_e$ score executed states, and let $u+v$ be a finite edit of the request. Its score response is
\begin{equation}
\delta_v\bar S_e(u)=s_e\!\left(T_e(u+v)\right)-s_e\!\left(T_e(u)\right).
\label{eq:transformation}
\end{equation}
A substantial change in requested motion need not produce a substantial score response: $T_e$ can attenuate it before measurement, and $h_e$ can flatten the remaining response. Section~\ref{sec:interventions} compares requested and executed motion and measures Eq.~\ref{eq:transformation} for smoothing and retiming interventions. These checks locate weak sensitivity in the complete composition; they do not separately identify each stage's contribution.

\paragraph{Closed-loop execution and gain transfer.}
\label{sec:regime}
Actions change subsequent planner inputs \citep{ross2011reduction}. Let $z_t$ contain observations, motion state, and history; $\Phi_e$ includes execution, state extraction, rendering, and history update. For a planner $\pi$,
\begin{equation}
u_t=\pi(z_t),\qquad z_{t+1}=\Phi_e(z_t,u_t).
\label{eq:feedback-system}
\end{equation}
Changing execution can alter subsequent requests even when initial predictions match. For $J_e(\pi)=\mathbb E_{P_{\pi,e}}[S_e]$, compare both planners under settings $a$ and $b$ on shared initial scenes:
\begin{equation}
I_{a,b}=\Delta S_a-\Delta S_b
=\bigl[J_a(\pi_R)-J_b(\pi_R)\bigr]-\bigl[J_a(\pi_0)-J_b(\pi_0)\bigr].
\label{eq:transfer-gap}
\end{equation}
Section~\ref{sec:closedloop} measures this interaction for two execution interfaces within otherwise fixed closed-loop conditions. A reversal requires opposite signs of $\Delta S_a$ and $\Delta S_b$; $I_{a,b}\ne0$ alone does not establish one. Comparing successive plans at the same future time then examines how the changed execution and feedback process affects later requests. Together, these measurements test gain transfer and describe the accompanying behavioral response, without assigning the reversal to a particular filtered motion component.

\label{sec:validation_requirements}
\label{sec:stress_test}
Score insensitivity permits a mismatch; $R$ and planner constraints also shape behavior. We therefore examine controlled interventions and optimized outputs separately, including cases where score gains and behavioral improvements agree.

\section{Behavioral Sensitivity of the Scoring Process}
\label{sec:stress-testing}

\subsection{Experimental Setup}
\label{sec:setup}
\label{sec:background}

\paragraph{Scores and targets.}
\label{sec:pdms}
\label{sec:score_training}
NAVSIM aggregates simulated subscores in $[0,1]$ using the original PDM Score notation \citep{dauner2024navsim}:
\begin{equation}
\textrm{PDMS} = \underbrace{\Bigg( { \prod_{m \in \{\texttt{NC}, \texttt{DAC}\}}} \texttt{score}_m \Bigg)}_{\text{penalties}} \times \underbrace{\Bigg( \frac{\sum_{w \in \{\texttt{EP}, \texttt{TTC}, \texttt{C}\}} \texttt{weight}_w \times \texttt{score}_w}{\sum_{w \in \{\texttt{EP}, \texttt{TTC}, \texttt{C}\}} \texttt{weight}_w } \Bigg)}_{\text{weighted average}}.
\label{eq:pdms}
\end{equation}
NC/DAC penalize at-fault collisions/road departures; EP/TTC/C measure progress, time-to-collision, and comfort with weights $5/5/2$. NC assigns $0.5$ for static-object collisions. DDC has zero aggregation weight. We use official scoring.
NAVSIM-v2 introduces EPDMS, with driving-direction compliance as a gate and additional lane-keeping and comfort components \citep{cao2025pseudo}.
We distinguish rescoring fixed outputs with EPDMS from the official two-stage protocol, which requires new predictions at original and synthetic states.

\paragraph{Planner comparisons.}
\label{sec:checkpoints}
ReCogDrive-2B/8B, Qwen-Drive, and NoRD provide within-architecture imitation-to-reinforcement comparisons \citep{li2026recogdrive,qwendrive2026,nord2026}.
ReCogDrive and NoRD use PDMS rewards; NoRD retains a fixed human-derived trajectory vocabulary.
Qwen-Drive uses EP/TTC/comfort weights $6/4/2$ and a displacement reward toward the logged trajectory.
GTRS-Dense changes expert supervision to reward-score distillation \citep{li2025gtrs}; TOAD adds fixed-weight search against predicted PDMS with comfort and anchoring penalties \citep{xu2026toad}.
DiffusionDrive v1 (IL)$\to$v2 (RL) evaluates the released version upgrade \citep{liao2025diffusiondrive,zou2025diffusiondrivev2}.
Released checkpoints are evaluated without retraining; Table~\ref{tab:pairs} reports absolute results and paired changes.

\paragraph{Evaluation.}
\label{sec:protocols}
\label{sec:behavior_metrics}
NAVSIM open-loop evaluation tracks a four-second plan with LQR (linear quadratic regulator) and a kinematic bicycle model while other agents replay their logs.
We use NAVSIM-v1 \texttt{navtest}, the \texttt{navhard} test set of NAVSIM-v2, and the WorldEngine test set (corner cases identified through failures of its VADv2-based planner on \texttt{navtest}) \citep{worldengine}.
WorldEngine supplies updated observations for replanning every 0.5\,s over four seconds. Our primary interface comparison uses all scenes with NR traffic, comparing the released trajectory-playback interface against LQR+bicycle execution under the same scoring settings. Earlier playback comparisons with NR/R traffic are reported separately in Appendix~\ref{app:closedloop}.
On nuPlan, Plan-R1 supplies an imitation/RL pair evaluated on \texttt{val14} scenarios from logs with non-reactive (NR) and reactive (R) traffic \citep{tang2026planr1,caesar2021nuplan}.
\label{sec:nuplan_setup}
Its reward changes CLS weights and gates.
Paired changes use scene and log bootstrap intervals and Holm-adjusted tests over fixed families. Evaluation protocols, planner configurations, and statistical procedures are detailed in Appendix~\ref{app:protocols}.

\subsection{Omission: Unscored Direction}
\label{sec:blindspots}

\paragraph{Constructing the blind spot.}
The Predictive Driver Model Score (PDMS) assigns zero aggregation weight to driving-direction compliance (DDC).
With the other scoring inputs fixed, $\partial\mathrm{PDMS}/\partial\mathrm{score}_{\mathrm{DDC}}=0$.
We test this invariance by laterally shifting route-following trajectories while keeping their speed profiles fixed and verifying lane direction against the map.
The 42 matched pairs across 37 \texttt{navhard} scenes pass the no at-fault collision (NC) and drivable-area compliance (DAC) checks, with near-equal progress and identical time-to-collision (TTC) and comfort scores.
The wrong-way trajectories travel 8.80\,m against the verified direction on average, yet the mean PDMS difference is only $1.9\times10^{-5}$ (Appendix~\ref{app:wrongway}).
Matching the other scored properties leaves a direction violation that the aggregate cannot distinguish.

Direction compliance also falls in optimized checkpoints whose aggregate scores improve.
Both ReCogDrive reinforcement learning (RL) checkpoints score higher than their imitation learning (IL) counterparts while direction compliance decreases (Table~\ref{tab:pairs}).
NoRD improves both its score and most measured diagnostics; its lower-scoring imitation baseline leaves greater headroom for improvement.
The omitted component can therefore deteriorate without directly lowering the score, but deterioration is not inevitable.

Test-time search exposes the same unconstrained dimension with fixed model weights.
In TOAD's cross-entropy method (CEM) search, the highest-predicted-score candidate gains official PDMS while direction compliance and the extended PDM score (EPDMS) fall (Figure~\ref{fig:toadsweep}).
TOAD's own iteration ablation also reports EPDMS deterioration with longer search \citep[Figure~4(c)]{xu2026toad}; our component analysis shows that direction compliance can fall while the v1 score rises.
The regularized deployment rule selects trajectories with higher direction compliance than the highest-predicted-score candidate later in the search.
Under the released v2 objective, which includes direction compliance, deployed mean compliance increases slightly on both original and synthetic scenes.
The unscored direction dimension worsens in both optimized ReCogDrive checkpoints and score-seeking TOAD candidates, while NoRD provides a counterexample.

\begin{table}[!t]
\caption{\textbf{NAVSIM \texttt{navtest}: scores and behavioral changes.}
Absolute values with within-family changes below optimized entries, computed before rounding; green/red denotes improvement/deterioration. EPDMS rescores fixed outputs.
DDC$^{\mathrm{v2}}$ excludes intersection positions.
Jitter: rad; Smooth./Adh.: smoothing displacement/plan-adherence error [m]; LK: lane keeping; HC: history comfort.
$^{\dagger}$ marks changes not supported by the paired statistical criteria in Appendix~\ref{app:uncertainty}.}

\label{tab:pairs}
\label{tab:v2pairs-main}
\centering\tablelayout
\setlength{\tabcolsep}{1.6pt}
\begin{tabular*}{\linewidth}{@{\extracolsep{\fill}}ccccccccccc@{}}
\toprule
& & \multicolumn{2}{c}{Aggregate scores} & \multicolumn{3}{c}{EPDMS components} & \multicolumn{4}{c}{Behavioral diagnostics} \\
\cmidrule(lr){3-4}\cmidrule(lr){5-7}\cmidrule(lr){8-11}
Planner & Variant & \metrichead{PDMS}{\uparrow} & \metrichead{EPDMS}{\uparrow} & \metrichead{DDC$^{\mathrm{v2}}$}{\uparrow} & \metrichead{LK}{\uparrow} & \metrichead{HC}{\uparrow} & \metrichead{DDC$^{\mathrm{v1}}$}{\uparrow} & \metrichead{Jitter}{\downarrow} & \metrichead{Smooth.}{\downarrow} & \metrichead{Adh.}{\downarrow} \\
\midrule
\parbox[c]{45pt}{\centering Ego-status\\MLP} & IL & $0.664$ & $0.645$ & $0.927$ & $0.792$ & $0.983$ & $0.904$ & $0.49$ & $0.086$ & $0.587$ \\
\addlinespace[4pt]
TransFuser & IL & $0.839$ & $0.814$ & $0.992$ & $0.848$ & $0.983$ & $0.980$ & $0.50$ & $0.092$ & $0.603$ \\
\addlinespace[4pt]
\parbox[c]{45pt}{\centering Latent\\TransFuser} & IL & $0.835$ & $0.810$ & $0.992$ & $0.853$ & $0.983$ & $0.979$ & $0.53$ & $0.091$ & $0.624$ \\
\midrule
\multirow[t]{2}{45pt}{\centering ReCogDrive\\2B} & IL & $0.863$ & $0.834$ & $0.993$ & $0.860$ & $0.983$ & $0.980$ & $0.50$ & $0.105$ & $0.575$ \\
 & \shortstack[c]{RL\\\strut} & \shortstack[c]{$0.905$\\{\scriptsize\textcolor{tableimprove}{\textbf{(+0.042)}}}} & \shortstack[c]{$0.854$\\{\scriptsize\textcolor{tableimprove}{\textbf{(+0.019)}}}} & \shortstack[c]{$0.982$\\{\scriptsize\textcolor{tabledegrade}{\textbf{(\textminus{}0.011)}}}} & \shortstack[c]{$0.802$\\{\scriptsize\textcolor{tabledegrade}{\textbf{(\textminus{}0.058)}}}} & \shortstack[c]{$0.977$\\{\scriptsize\textcolor{tabledegrade}{\textbf{(\textminus{}0.006)}}}} & \shortstack[c]{$0.967$\\{\scriptsize\textcolor{tabledegrade}{\textbf{(\textminus{}0.013)}}}} & \shortstack[c]{$1.69$\\{\scriptsize\textcolor{tabledegrade}{\textbf{(+1.19)}}}} & \shortstack[c]{$0.288$\\{\scriptsize\textcolor{tabledegrade}{\textbf{(+0.183)}}}} & \shortstack[c]{$0.893$\\{\scriptsize\textcolor{tabledegrade}{\textbf{(+0.319)}}}} \\
\addlinespace[3pt]
\multirow[t]{2}{45pt}{\centering ReCogDrive\\8B} & IL & $0.869$ & $0.842$ & $0.994$ & $0.857$ & $0.983$ & $0.980$ & $0.64$ & $0.109$ & $0.583$ \\
 & \shortstack[c]{RL\\\strut} & \shortstack[c]{$0.903$\\{\scriptsize\textcolor{tableimprove}{\textbf{(+0.035)}}}} & \shortstack[c]{$0.859$\\{\scriptsize\textcolor{tableimprove}{\textbf{(+0.017)}}}} & \shortstack[c]{$0.989$\\{\scriptsize\textcolor{tabledegrade}{\textbf{(\textminus{}0.005)}}}} & \shortstack[c]{$0.832$\\{\scriptsize\textcolor{tabledegrade}{\textbf{(\textminus{}0.025)}}}} & \shortstack[c]{$0.979$\\{\scriptsize\textcolor{tabledegrade}{\textbf{(\textminus{}0.004)}}}} & \shortstack[c]{$0.974$\\{\scriptsize\textcolor{tabledegrade}{\textbf{(\textminus{}0.006)}}}} & \shortstack[c]{$1.56$\\{\scriptsize\textcolor{tabledegrade}{\textbf{(+0.92)}}}} & \shortstack[c]{$0.281$\\{\scriptsize\textcolor{tabledegrade}{\textbf{(+0.173)}}}} & \shortstack[c]{$0.982$\\{\scriptsize\textcolor{tabledegrade}{\textbf{(+0.399)}}}} \\
\addlinespace[3pt]
Qwen-Drive & IL & $0.879$ & $0.854$ & $0.996$ & $0.863$ & $0.982$ & $0.987$ & $0.48$ & $0.116$ & $0.609$ \\
 & \shortstack[c]{RL\\\strut} & \shortstack[c]{$0.905$\\{\scriptsize\textcolor{tableimprove}{\textbf{(+0.026)}}}} & \shortstack[c]{$0.873$\\{\scriptsize\textcolor{tableimprove}{\textbf{(+0.019)}}}} & \shortstack[c]{$0.995$\\{\scriptsize\textcolor{tabledegrade}{\textbf{(\textminus{}0.001)}}}} & \shortstack[c]{$0.851$\\{\scriptsize\textcolor{tabledegrade}{\textbf{(\textminus{}0.012)}}}} & \shortstack[c]{$0.978$\\{\scriptsize\textcolor{tabledegrade}{\textbf{(\textminus{}0.004)}}}} & \shortstack[c]{$0.985$\\{\scriptsize\textcolor{tabledegrade}{\textbf{(\textminus{}0.001\textsuperscript{$\dagger$})}}}} & \shortstack[c]{$0.55$\\{\scriptsize\textcolor{tabledegrade}{\textbf{(+0.07)}}}} & \shortstack[c]{$0.212$\\{\scriptsize\textcolor{tabledegrade}{\textbf{(+0.097)}}}} & \shortstack[c]{$0.564$\\{\scriptsize\textcolor{tableimprove}{\textbf{(\textminus{}0.046)}}}} \\
\addlinespace[3pt]
\multirow[t]{2}{45pt}{\centering Diffusion\\Drive} & v1 IL & $0.881$ & $0.852$ & $0.994$ & $0.860$ & $0.984$ & $0.981$ & $0.62$ & $0.103$ & $0.584$ \\
 & \shortstack[c]{v2 RL\\\strut} & \shortstack[c]{$0.909$\\{\scriptsize\textcolor{tableimprove}{\textbf{(+0.029)}}}} & \shortstack[c]{$0.855$\\{\scriptsize\textcolor{tableimprove}{\textbf{(+0.003\textsuperscript{$\dagger$})}}}} & \shortstack[c]{$0.985$\\{\scriptsize\textcolor{tabledegrade}{\textbf{(\textminus{}0.009)}}}} & \shortstack[c]{$0.840$\\{\scriptsize\textcolor{tabledegrade}{\textbf{(\textminus{}0.020)}}}} & \shortstack[c]{$0.891$\\{\scriptsize\textcolor{tabledegrade}{\textbf{(\textminus{}0.093)}}}} & \shortstack[c]{$0.968$\\{\scriptsize\textcolor{tabledegrade}{\textbf{(\textminus{}0.013)}}}} & \shortstack[c]{$1.99$\\{\scriptsize\textcolor{tabledegrade}{\textbf{(+1.37)}}}} & \shortstack[c]{$0.387$\\{\scriptsize\textcolor{tabledegrade}{\textbf{(+0.284)}}}} & \shortstack[c]{$1.240$\\{\scriptsize\textcolor{tabledegrade}{\textbf{(+0.656)}}}} \\
\addlinespace[3pt]
NoRD & IL & $0.738$ & $0.715$ & $0.967$ & $0.813$ & $0.957$ & $0.946$ & $0.99$ & $0.157$ & $0.743$ \\
 & \shortstack[c]{RL\\\strut} & \shortstack[c]{$0.846$\\{\scriptsize\textcolor{tableimprove}{\textbf{(+0.108)}}}} & \shortstack[c]{$0.812$\\{\scriptsize\textcolor{tableimprove}{\textbf{(+0.097)}}}} & \shortstack[c]{$0.982$\\{\scriptsize\textcolor{tableimprove}{\textbf{(+0.015)}}}} & \shortstack[c]{$0.827$\\{\scriptsize\textcolor{tableimprove}{\textbf{(+0.014)}}}} & \shortstack[c]{$0.977$\\{\scriptsize\textcolor{tableimprove}{\textbf{(+0.021)}}}} & \shortstack[c]{$0.963$\\{\scriptsize\textcolor{tableimprove}{\textbf{(+0.017)}}}} & \shortstack[c]{$0.78$\\{\scriptsize\textcolor{tableimprove}{\textbf{(\textminus{}0.22)}}}} & \shortstack[c]{$0.126$\\{\scriptsize\textcolor{tableimprove}{\textbf{(\textminus{}0.030)}}}} & \shortstack[c]{$0.745$\\{\scriptsize\textcolor{tabledegrade}{\textbf{(+0.003\textsuperscript{$\dagger$})}}}} \\
\addlinespace[3pt]
\multirow[t]{2}{45pt}{\centering GTRS\\Dense} & Expert & $0.893$ & $0.876$ & $0.994$ & $0.875$ & $0.969$ & $0.985$ & $0.48$ & $0.088$ & $1.162$ \\
 & \shortstack[c]{Distill\\\strut} & \shortstack[c]{$0.903$\\{\scriptsize\textcolor{tableimprove}{\textbf{(+0.010)}}}} & \shortstack[c]{$0.879$\\{\scriptsize\textcolor{tableimprove}{\textbf{(+0.003\textsuperscript{$\dagger$})}}}} & \shortstack[c]{$0.993$\\{\scriptsize\textcolor{tabledegrade}{\textbf{(\textminus{}0.001\textsuperscript{$\dagger$})}}}} & \shortstack[c]{$0.866$\\{\scriptsize\textcolor{tabledegrade}{\textbf{(\textminus{}0.009)}}}} & \shortstack[c]{$0.977$\\{\scriptsize\textcolor{tableimprove}{\textbf{(+0.008)}}}} & \shortstack[c]{$0.981$\\{\scriptsize\textcolor{tabledegrade}{\textbf{(\textminus{}0.004)}}}} & \shortstack[c]{$0.50$\\{\scriptsize\textcolor{tabledegrade}{\textbf{(+0.02\textsuperscript{$\dagger$})}}}} & \shortstack[c]{$0.087$\\{\scriptsize\textcolor{tableimprove}{\textbf{(\textminus{}0.001)}}}} & \shortstack[c]{$1.112$\\{\scriptsize\textcolor{tableimprove}{\textbf{(\textminus{}0.050)}}}} \\
\addlinespace[3pt]
\multirow[t]{2}{45pt}{\centering DrivoR /\\TOAD} & Base & $0.948$ & $0.893$ & $0.985$ & $0.849$ & $0.967$ & $0.969$ & $0.39$ & $0.127$ & $0.708$ \\
 & \shortstack[c]{CEM\\\strut} & \shortstack[c]{$0.949$\\{\scriptsize\textcolor{tableimprove}{\textbf{(+0.001\textsuperscript{$\dagger$})}}}} & \shortstack[c]{$0.890$\\{\scriptsize\textcolor{tabledegrade}{\textbf{(\textminus{}0.003)}}}} & \shortstack[c]{$0.982$\\{\scriptsize\textcolor{tabledegrade}{\textbf{(\textminus{}0.003)}}}} & \shortstack[c]{$0.840$\\{\scriptsize\textcolor{tabledegrade}{\textbf{(\textminus{}0.008)}}}} & \shortstack[c]{$0.967$\\{\scriptsize\textcolor{tabledegrade}{\textbf{(\textminus{}0.001\textsuperscript{$\dagger$})}}}} & \shortstack[c]{$0.965$\\{\scriptsize\textcolor{tabledegrade}{\textbf{(\textminus{}0.004)}}}} & \shortstack[c]{$0.25$\\{\scriptsize\textcolor{tableimprove}{\textbf{(\textminus{}0.14)}}}} & \shortstack[c]{$0.129$\\{\scriptsize\textcolor{tabledegrade}{\textbf{(+0.002)}}}} & \shortstack[c]{$0.716$\\{\scriptsize\textcolor{tabledegrade}{\textbf{(+0.008)}}}} \\
\bottomrule
\end{tabular*}
\end{table}

\subsection{Thresholding: Motion within Passing Regions}
\label{sec:nuplan}

\paragraph{Passing regions and score ties.}
The flat regions in Eq.~\ref{eq:flat-region} predict that a passing subscore can conceal continuous behavioral changes; we test its resolution through motion diagnostics and fixed-output rescoring.
Thresholded comfort leaves continuous motion differences unresolved within the passing region.
Across the evaluated NAVSIM configurations, comfort fails in only 0.14\% of planner--scene evaluations.
For DiffusionDrive v1, v2, and Human on \texttt{navhard}, even the largest within-scene threshold utilization averages roughly one-third of the bound (Appendix Figure~\ref{fig:absorbed}).

At the aggregate level, score ties provide a complementary check of behavioral resolution during trajectory selection.
In DiffusionDrive v2's candidate pools, 274 of 450 \texttt{navhard} scenes have multiple PDMS maximizers with different heading jitter.
The aggregate provides no preference among those geometrically different plans.

\paragraph{Changes after optimization.}
We further examine in nuPlan whether improved closed-loop task scores are accompanied by deterioration in continuous measures of executed motion.
We measure the largest magnitude of longitudinal jerk, the time derivative of longitudinal acceleration:
\begin{equation}
D_{\mathrm{jerk}}=\max_t|j_\parallel(t)|,\qquad
j_\parallel(t)=\frac{\mathrm d a_\parallel(t)}{\mathrm dt}.
\label{eq:jerk-diagnostic}
\end{equation}
Plan-R1's mean absolute jerk peak increases after RL in both non-reactive and reactive traffic, with both changes passing the paired statistical checks.
Its closed-loop score (CLS), road compliance, progress, and speed-limit compliance improve, while planned jitter and plan churn decrease (Table~\ref{tab:nuplan}).
The binary comfort term has no supported mean change in non-reactive traffic and decreases slightly in reactive traffic.
The continuous measurement identifies a motion change that the comfort pass rate alone does not describe.

\begin{table}[!t]
\centering
\caption{\textbf{nuPlan \texttt{val14}.} Plan-R1 IL$\to$RL under non-reactive (NR) and reactive (R) traffic.}
\label{tab:nuplan}
\tablelayout
\begin{tabular*}{\linewidth}{@{\extracolsep{\fill}}llrr@{}}
\toprule
Quantity & Role in CLS & $\Delta$ NR & $\Delta$ R \\
\midrule
CLS $\uparrow$ & Aggregate & \textcolor{tableimprove}{$\mathbf{+0.034}$} & \textcolor{tableimprove}{$\mathbf{+0.067}$} \\
Drivable-area compliance $\uparrow$ & Gate & \textcolor{tableimprove}{$\mathbf{+0.042}$} & \textcolor{tableimprove}{$\mathbf{+0.033}$} \\
Planned heading jitter $\downarrow$ & Absent & \textcolor{tableimprove}{$\mathbf{-0.235}$} & \textcolor{tableimprove}{$\mathbf{-0.538}$} \\
Absolute lon.\ jerk peak [m/s$^3$] $\downarrow$ & Threshold & \textcolor{tabledegrade}{$\mathbf{+0.136}$} & \textcolor{tabledegrade}{$\mathbf{+0.206}$} \\
Absolute jerk peak [m/s$^3$] $\downarrow$ & Threshold & \textcolor{tabledegrade}{$\mathbf{+0.068}$} & \textcolor{tabledegrade}{$\mathbf{+0.145}$} \\
Comfort term $\uparrow$ & Weighted term & \textcolor{tabledegrade}{$\mathbf{-0.002^{\dagger}}$} & \textcolor{tabledegrade}{$\mathbf{-0.009}$} \\
\bottomrule
\end{tabular*}
\end{table}

\paragraph{Changing score resolution.}
A continuous comfort score can distinguish motion within the passing region, but it still evaluates motion after tracking.
We replace the binary mapping $h_e$ with three continuous alternatives while keeping the execution transformation $T_e$ and planner outputs fixed.
All three preserve DiffusionDrive v2's score advantage on both \texttt{navhard} and the WorldEngine test set (Appendix Table~\ref{tab:continuous}).

\subsection{Execution Transformation: Requested versus Scored Motion}
\label{sec:pairs}
\label{sec:pace}
\label{sec:interventions}

In Section~\ref{sec:transformation}, we predict that substantial request changes can receive weak score responses after execution. we test this through motion-layer comparisons and finite trajectory interventions (Eq.~\ref{eq:transformation}).
Smoothing displacement measures how far waypoints move under a local smoothing operation; heading jitter measures variation in successive heading changes.
For eight planned positions $p_1,\ldots,p_N$ spaced 0.5\,s apart, one three-point moving average keeps the endpoints fixed and sets $\tilde p_i=(p_{i-1}+p_i+p_{i+1})/3$ for internal points.
The two diagnostics are
\begin{equation}
D_{\mathrm{smooth}}=\max_i\|\tilde p_i-p_i\|_2,
\qquad
D_{\mathrm{jitter}}=\sum_{i=2}^{N-1}|\theta_{i+1}-2\theta_i+\theta_{i-1}|,
\label{eq:plan-geometry}
\end{equation}
where $\theta_i$ is the unwrapped heading of segment $p_{i-1}\to p_i$, with $p_0$ the ego origin.
Smoothing displacement captures both path bending and waypoint spacing; heading jitter is interpreted with path length.
Plan-adherence error measures the maximum separation between the interpolated request $p^{\mathrm{plan}}(t)$ and the tracked position $\hat p(t)$:
\begin{equation}
D_{\mathrm{adh}}=\max_t\|\hat p(t)-p^{\mathrm{plan}}(t)\|_2.
\label{eq:adherence}
\end{equation}

\paragraph{Attenuation through tracking.}
Planned and executed motion show different changes after optimization.
For ReCogDrive-2B, the mean planned absolute acceleration peak rises from 1.14 to 2.84\,m/s$^2$ after RL, while the executed longitudinal peak changes from 0.88 to 1.20\,m/s$^2$.
The accompanying increase in plan-adherence error records greater separation between the requested and tracked trajectories (Table~\ref{tab:pairs}).
For Qwen-Drive RL, planned acceleration/deceleration peaks are 1.65/1.88\,m/s$^2$, whereas the simulated peaks are 0.28/0.53 (Table~\ref{tab:qwen-pace}). These peaks describe different motion layers and do not directly measure how much of a request is discarded.

\begin{figure}[!t]
\centering
\includegraphics[width=\linewidth]{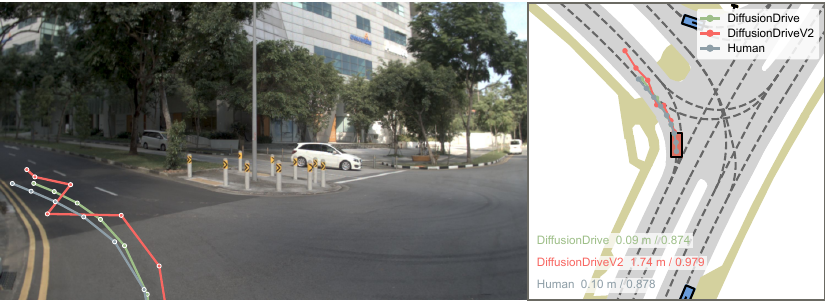}
\caption{\textbf{Illustrative trajectory comparison.} WorldEngine case; labels: smoothing displacement (m, lower is better) / PDMS. The oscillatory v2 plan scores 0.979 versus v1 0.874 and Human 0.878.}
\label{fig:qualitative-selected}
\end{figure}

\paragraph{Finite interventions.}
Large changes to requested geometry produce little response from the complete execution--scoring pipeline.
Smoothing DiffusionDrive v2's positions while preserving its output headings removes 70--74\% of heading jitter on \texttt{navhard} and the WorldEngine test set, with mean PDMS changes below 0.01 in magnitude.
The illustrative case shows a visibly oscillatory v2 plan scoring above both v1 and Human, although all three pass comfort after tracking (Figure~\ref{fig:qualitative-selected}).

Qwen-Drive's increased smoothing displacement is concentrated in longitudinal waypoint spacing, accompanying changes in planned speed; its plan-adherence error improves.
Retiming along each original path preserves endpoint speeds and total length while reducing the RL plan's peak speed change per unit time by nearly half, while mean PDMS changes by only $-0.0002$.
Both interventions alter requests before tracking, so their score response includes the execution transformation.

\paragraph{Contrasting responses.}
NoRD improves PDMS while reducing jitter and smoothing displacement; GTRS-Dense improves adherence, with little change in smoothing displacement.
TOAD reduces jitter while slightly increasing smoothing displacement and adherence error.
Requested-motion diagnostics distinguish plan quality from the controller's ability to accommodate requests; execution scores alone cannot establish both.

\paragraph{Interpreting aggregate gains.}\label{sec:aggregation-evidence}
The three blind spots also motivate separating behavioral changes from how the aggregate credits them: reported progress can rise through more trajectories passing scoring gates.
PDMS assigns zero progress outside $A=\{\mathrm{NC}\cdot\mathrm{DAC}>0\}$, so
\begin{equation}
\mathbb E[\mathrm{EP}_{\mathrm{rep}}]
=\Pr(A)\,\mathbb E[\mathrm{EP}_{\mathrm{rep}}\mid A].
\label{eq:ep-decomposition}
\end{equation}
For NoRD, the pass rate rises by about 11 percentage points while conditional progress remains nearly unchanged, increasing reported progress despite slower, shorter plans.
Qwen-Drive improves both the pass rate and conditional progress.
Interpreting a progress gain therefore requires separating these two contributions.

\begin{table}[!tp]
\centering
\caption{\textbf{NAVSIM-v2 \texttt{navhard-two-stage} results.}
Selected EPDMS components on original (Stage 1) and synthetic (Stage 2) scenes, and combined score. RL entries show absolute values and parenthesized RL-minus-IL changes. EC: two-frame extended comfort.}
\label{tab:twostage-main}
\tablelayout
\setlength{\tabcolsep}{2.2pt}
\begin{tabular*}{\linewidth}{@{\extracolsep{\fill}}ccccccccccc@{}}
\toprule
& & \multicolumn{4}{c}{Stage 1 components} & \multicolumn{4}{c}{Stage 2 components} & Overall \\
\cmidrule(lr){3-6}\cmidrule(lr){7-10}\cmidrule(lr){11-11}
Planner & Variant & \metrichead{DAC}{\uparrow} & \metrichead{DDC}{\uparrow} & \metrichead{EP}{\uparrow} & \metrichead{EC}{\uparrow} & \metrichead{DAC}{\uparrow} & \metrichead{DDC}{\uparrow} & \metrichead{EP}{\uparrow} & \metrichead{EC}{\uparrow} & \metrichead{EPDMS}{\uparrow} \\
\midrule
\multirow[t]{2}{45pt}{\centering ReCogDrive\\2B} & IL & $0.827$ & $0.984$ & $0.828$ & $0.764$ & $0.721$ & $0.823$ & $0.792$ & $0.688$ & $0.278$ \\
 & \shortstack[c]{RL\\\strut} & \shortstack[c]{$0.907$\\{\scriptsize\textcolor{tableimprove}{\textbf{(+0.080)}}}} & \shortstack[c]{$0.956$\\{\scriptsize\textcolor{tabledegrade}{\textbf{(\textminus{}0.029)}}}} & \shortstack[c]{$0.856$\\{\scriptsize\textcolor{tableimprove}{\textbf{(+0.028)}}}} & \shortstack[c]{$0.329$\\{\scriptsize\textcolor{tabledegrade}{\textbf{(\textminus{}0.436)}}}} & \shortstack[c]{$0.726$\\{\scriptsize\textcolor{tableimprove}{\textbf{(+0.005)}}}} & \shortstack[c]{$0.812$\\{\scriptsize\textcolor{tabledegrade}{\textbf{(\textminus{}0.011)}}}} & \shortstack[c]{$0.853$\\{\scriptsize\textcolor{tableimprove}{\textbf{(+0.061)}}}} & \shortstack[c]{$0.310$\\{\scriptsize\textcolor{tabledegrade}{\textbf{(\textminus{}0.378)}}}} & \shortstack[c]{$0.270$\\{\scriptsize\textcolor{tabledegrade}{\textbf{(\textminus{}0.008\textsuperscript{$\dagger$})}}}} \\
\addlinespace[3pt]
\multirow[t]{2}{45pt}{\centering ReCogDrive\\8B} & IL & $0.842$ & $0.990$ & $0.834$ & $0.769$ & $0.721$ & $0.821$ & $0.808$ & $0.680$ & $0.263$ \\
 & \shortstack[c]{RL\\\strut} & \shortstack[c]{$0.911$\\{\scriptsize\textcolor{tableimprove}{\textbf{(+0.069)}}}} & \shortstack[c]{$0.973$\\{\scriptsize\textcolor{tabledegrade}{\textbf{(\textminus{}0.017)}}}} & \shortstack[c]{$0.860$\\{\scriptsize\textcolor{tableimprove}{\textbf{(+0.027)}}}} & \shortstack[c]{$0.204$\\{\scriptsize\textcolor{tabledegrade}{\textbf{(\textminus{}0.564)}}}} & \shortstack[c]{$0.728$\\{\scriptsize\textcolor{tableimprove}{\textbf{(+0.008)}}}} & \shortstack[c]{$0.819$\\{\scriptsize\textcolor{tabledegrade}{\textbf{(\textminus{}0.002)}}}} & \shortstack[c]{$0.864$\\{\scriptsize\textcolor{tableimprove}{\textbf{(+0.056)}}}} & \shortstack[c]{$0.153$\\{\scriptsize\textcolor{tabledegrade}{\textbf{(\textminus{}0.527)}}}} & \shortstack[c]{$0.255$\\{\scriptsize\textcolor{tabledegrade}{\textbf{(\textminus{}0.008\textsuperscript{$\dagger$})}}}} \\
\addlinespace[3pt]
\multirow[t]{2}{45pt}{\centering Qwen-Drive} & IL & $0.887$ & $0.994$ & $0.829$ & $0.813$ & $0.753$ & $0.843$ & $0.771$ & $0.647$ & $0.303$ \\
 & \shortstack[c]{RL\\\strut} & \shortstack[c]{$0.940$\\{\scriptsize\textcolor{tableimprove}{\textbf{(+0.053)}}}} & \shortstack[c]{$0.997$\\{\scriptsize\textcolor{tableimprove}{\textbf{(+0.002)}}}} & \shortstack[c]{$0.839$\\{\scriptsize\textcolor{tableimprove}{\textbf{(+0.010)}}}} & \shortstack[c]{$0.342$\\{\scriptsize\textcolor{tabledegrade}{\textbf{(\textminus{}0.471)}}}} & \shortstack[c]{$0.782$\\{\scriptsize\textcolor{tableimprove}{\textbf{(+0.029)}}}} & \shortstack[c]{$0.851$\\{\scriptsize\textcolor{tableimprove}{\textbf{(+0.008)}}}} & \shortstack[c]{$0.777$\\{\scriptsize\textcolor{tableimprove}{\textbf{(+0.006)}}}} & \shortstack[c]{$0.351$\\{\scriptsize\textcolor{tabledegrade}{\textbf{(\textminus{}0.296)}}}} & \shortstack[c]{$0.306$\\{\scriptsize\textcolor{tableimprove}{\textbf{(+0.003\textsuperscript{$\dagger$})}}}} \\
\addlinespace[3pt]
\multirow[t]{2}{45pt}{\centering NoRD} & IL & $0.651$ & $0.923$ & $0.810$ & $0.604$ & $0.593$ & $0.769$ & $0.767$ & $0.525$ & $0.178$ \\
 & \shortstack[c]{RL\\\strut} & \shortstack[c]{$0.836$\\{\scriptsize\textcolor{tableimprove}{\textbf{(+0.184)}}}} & \shortstack[c]{$0.973$\\{\scriptsize\textcolor{tableimprove}{\textbf{(+0.050)}}}} & \shortstack[c]{$0.811$\\{\scriptsize\textcolor{tableimprove}{\textbf{(+0.001)}}}} & \shortstack[c]{$0.653$\\{\scriptsize\textcolor{tableimprove}{\textbf{(+0.049)}}}} & \shortstack[c]{$0.713$\\{\scriptsize\textcolor{tableimprove}{\textbf{(+0.120)}}}} & \shortstack[c]{$0.820$\\{\scriptsize\textcolor{tableimprove}{\textbf{(+0.051)}}}} & \shortstack[c]{$0.808$\\{\scriptsize\textcolor{tableimprove}{\textbf{(+0.040)}}}} & \shortstack[c]{$0.563$\\{\scriptsize\textcolor{tableimprove}{\textbf{(+0.038)}}}} & \shortstack[c]{$0.257$\\{\scriptsize\textcolor{tableimprove}{\textbf{(+0.079)}}}} \\
\bottomrule
\end{tabular*}
\end{table}

\paragraph{Broader coverage and paired-state evaluation.}
\label{sec:coverage-evidence}
Expanding what is scored exposes additional trade-offs but does not ensure that every behavior improves with the aggregate.
When fixed outputs are rescored with EPDMS, both ReCogDrive pairs retain score gains while lane keeping and history comfort decrease; DiffusionDrive v2's score advantage becomes much smaller, with an interval containing zero, while history comfort also decreases (Table~\ref{tab:pairs}).
TOAD's deployed search changes PDMS by $+0.001$ and EPDMS by $-0.003$, with the PDMS log interval including zero; NoRD improves all displayed EPDMS components.
The separate official two-stage evaluation generates predictions at original and synthetic states and measures extended comfort from time-aligned simulated motion in consecutive plans.
ReCogDrive and Qwen-Drive improve drivable-area compliance and progress in both stages while extended comfort falls, but the combined-score changes of all three have paired intervals containing zero; NoRD improves all displayed components and retains a positive combined gain (Table~\ref{tab:twostage-main}).
Broader coverage reveals these inconsistencies, while the preceding motion-layer tests leave a further question: whether gains survive a change in execution during repeated replanning.

\section{RL Gains Depend on the Execution Interface}
\label{sec:closedloop}

The benefit of optimization depends on the process that executes requests and returns them as feedback.
Section~\ref{sec:interventions} shows that better tracked scores can coexist with poorer requested-motion diagnostics and weak responses to substantial trajectory edits.
The closed-loop recurrence in Eq.~\ref{eq:feedback-system} adds another consequence: execution changes the inputs on which later requests depend.
We test whether gains persist across execution interfaces, then examine how successive predictions change during replanning.
We compare LQR with a bicycle model against playback on matched WorldEngine scenarios, keeping non-reactive traffic, scoring, rendering settings, and inference seeds fixed.
Initial predictions agree exactly across interfaces for each checkpoint.
This isolates the effect of execution and its subsequent feedback; protocol details and behavioral checks appear in Appendix~\ref{app:interface-control}.

\begin{table}[!ht]
\centering\tablelayout
\caption{\textbf{Gain reversals across execution interfaces.} Mean PDMS on matched NR scenes per family. Ref./Opt.: IL/base$\to$RL, except TOAD Frozen$\to$CEM; $\Delta$: Opt.$-$Ref. Interaction: $I_{\mathrm{LQR},\mathrm{PB}}=\Delta S_{\mathrm{LQR}}-\Delta S_{\mathrm{PB}}$ (Eq.~\ref{eq:transfer-gap}).}
\label{tab:execution-interface}
\scriptsize
\setlength{\tabcolsep}{1.5pt}
\renewcommand{\arraystretch}{1.0}
\begin{tabular*}{\linewidth}{@{\extracolsep{\fill}}lrrrrrrrrc@{}}
\toprule
& \multicolumn{3}{c}{LQR + bicycle model} & \multicolumn{3}{c}{Playback} & \multicolumn{2}{c}{LQR$-$PB} & Interaction \\
\cmidrule(lr){2-4}\cmidrule(lr){5-7}\cmidrule(lr){8-9}
Planner & Ref. & Opt. & $\Delta$ & Ref. & Opt. & $\Delta$ & Opt. & Ref. & $I_{\mathrm{LQR},\mathrm{PB}}$ [95\% CI] \\
\midrule
ReCogDrive-2B & 0.580 & 0.798 & \textcolor{tableimprove}{$\mathbf{+0.218}$} & 0.688 & 0.452 & \textcolor{tabledegrade}{$\mathbf{-0.236}$} & \textcolor{tableimprove}{$\mathbf{+0.346}$} & \textcolor{tabledegrade}{$\mathbf{-0.108}$} & {\scriptsize$\mathbf{+0.454}\;[+0.377,+0.533]$} \\
\addlinespace[1.5pt]
ReCogDrive-8B & 0.628 & 0.786 & \textcolor{tableimprove}{$\mathbf{+0.158}$} & 0.690 & 0.490 & \textcolor{tabledegrade}{$\mathbf{-0.200}$} & \textcolor{tableimprove}{$\mathbf{+0.295}$} & \textcolor{tabledegrade}{$\mathbf{-0.062}$} & {\scriptsize$\mathbf{+0.358}\;[+0.274,+0.436]$} \\
\addlinespace[1.5pt]
Qwen-Drive & 0.697 & 0.834 & \textcolor{tableimprove}{$\mathbf{+0.136}$} & 0.787 & 0.398 & \textcolor{tabledegrade}{$\mathbf{-0.388}$} & \textcolor{tableimprove}{$\mathbf{+0.435}$} & \textcolor{tabledegrade}{$\mathbf{-0.089}$} & {\scriptsize$\mathbf{+0.525}\;[+0.448,+0.603]$} \\
\midrule
NoRD & 0.462 & 0.648 & \textcolor{tableimprove}{$\mathbf{+0.186}$} & 0.516 & 0.672 & \textcolor{tableimprove}{$\mathbf{+0.156}$} & \textcolor{tabledegrade}{$\mathbf{-0.024}$} & \textcolor{tabledegrade}{$\mathbf{-0.054}$} & {\scriptsize$+0.030\;[-0.044,+0.102]$} \\
\addlinespace[1.5pt]
TOAD & 0.843 & 0.916 & \textcolor{tableimprove}{$\mathbf{+0.073}$} & 0.761 & 0.854 & \textcolor{tableimprove}{$\mathbf{+0.093}$} & \textcolor{tableimprove}{$\mathbf{+0.063}$} & \textcolor{tableimprove}{$\mathbf{+0.082}$} & {\scriptsize$-0.020\;[-0.062,+0.022]$} \\
\bottomrule
\end{tabular*}
\end{table}

For both ReCogDrive sizes and Qwen-Drive, switching from playback to LQR lowers the imitation baseline's score but raises the RL checkpoint's score (Table~\ref{tab:execution-interface}, LQR$-$PB columns).
Changing execution reverses the sign of the RL gain for both ReCogDrive sizes and Qwen-Drive (Table~\ref{tab:execution-interface}).
All three improve under LQR tracking and deteriorate under playback, with log-bootstrap intervals excluding zero in both directions; NoRD and TOAD retain positive gains under both.
The interaction intervals exclude zero for all three reversing pairs; Qwen-Drive's change in RL gain is $+0.525$.
The benefit of optimization therefore depends on the execution interface.
For Qwen-Drive, road-departure counts rise from 26 to 111 after RL under playback, but fall from 66 to 27 under LQR.
Under playback, its IL-to-RL drivable-area compliance falls from 0.91 to 0.61 and ego progress from 0.68 to 0.40 (Table~\ref{tab:interface-components}).

A single-plan control distinguishes executing a fixed request from repeated replanning.
In a separate Qwen-Drive control with the same first plan frozen, replacing LQR with playback changes the RL score by only $-0.005$, with a log-bootstrap 95\% interval of $[-0.035,+0.024]$.
This small contrast motivates examining subsequent requests and returned states; the controls are not an additive causal decomposition (Appendix~\ref{app:interface-control}).

Replanning can repeatedly replace a predicted slowdown with a higher speed request.
To distinguish these effects, let $\hat v_k(r)$ denote longitudinal speed at relative time $r$ in plan $k$, and let $h=0.5$\,s be the replanning interval.
Under playback, the input speed is $v_k=\hat v_{k-1}(h)$, giving, for $k\ge1$,
\begin{equation}
v_{k+1}-v_k
=\underbrace{\hat v_{k-1}(2h)-\hat v_{k-1}(h)}_{\text{previous plan's speed change}}
+\underbrace{\hat v_k(h)-\hat v_{k-1}(2h)}_{\text{same-time replanning revision}}.
\label{eq:speed-revision}
\end{equation}
The second term compares consecutive predictions for the same absolute future time, separating a change already intended by the previous plan from a revision introduced by replanning.

The earlier matched Qwen-Drive NR playback rollouts show a persistent contrast between these two quantities.
Of 2,296 RL plans, 1,897 (82.6\%) predict slowing down from $h$ to $2h$, yet all seven same-time revisions are positive in 196 RL scenes, compared with one IL scene.
Mean revisions are $+2.184$\,m/s for RL and $-0.262$\,m/s for IL; the saved records satisfy Eq.~\ref{eq:speed-revision} to numerical precision (Appendix~\ref{app:execution-feedback}).
They show how slowing within individual plans can coexist with repeated upward revisions during execution.

\begin{figure}[!t]
\centering
\includegraphics[width=\linewidth]{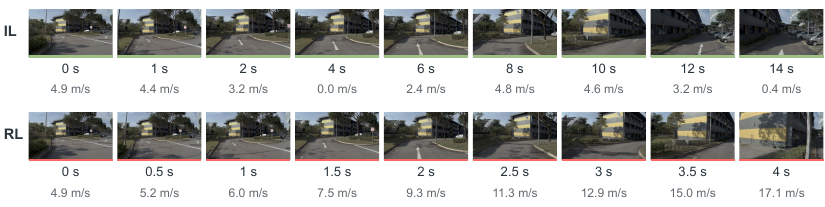}
\caption{\textbf{Qwen-Drive at the same junction.} Historical WorldEngine non-reactive playback front-camera renders with timestamps and longitudinal speeds.}
\label{fig:qwen-timeline}
\end{figure}

Successive predictions show that the execution interface also changes the planner's response to feedback.
On the full interface cohort, the same-time revision $\hat v_k(h)-\hat v_{k-1}(2h)$ averages $+2.17$\,m/s for RL under playback, versus $+0.58$ under LQR; the IL changes are $-0.27$ and $-0.06$\,m/s, respectively.
The paired change in RL's speed revision is $-1.58$\,m/s, with a log-bootstrap 95\% interval of $[-1.71,-1.45]$.
The matched intervention therefore changes subsequent requests as well as executed motion; the speed revisions describe this response without isolating its causal contribution to the score reversal.
In the earlier playback evaluation, Qwen-Drive's RL planner repeatedly replaces a predicted slowdown with a higher near-term speed and leaves the road (Figure~\ref{fig:qwen-timeline}).
The IL planner initially stops, then continues toward the same junction while remaining on road beyond the evaluation horizon.
ReCogDrive's earlier playback records likewise show more frequent predicted forward--backward motion after RL (Table~\ref{tab:feedback-foldbacks}).
These examples illustrate successive requests; the matched comparison establishes the gain reversal.

\paragraph{Interpretation.}
The matched results establish that RL gains can depend on the execution interface and its subsequent feedback.
One possible explanation is that tracking suppresses request variations whose consequences emerge under playback, but the contribution of specific filtered components remains to be isolated (Appendix~\ref{app:interface-control}).
An improvement claim therefore concerns both the planner's requests and the process that executes, scores, and returns them as feedback.

\section{Conclusions \& Take-aways}
\label{sec:conclusion}

Our results show that metric validity under optimization cannot be inferred from aggregate score gains alone.
Behavioral distinctions can be lost through omission, thresholding, and transformations between requested and executed motion.
Gains can also reverse across execution interfaces during repeated replanning, showing that the benefit of optimization depends on how requests are executed and returned as feedback.
These findings motivate evaluating both the behavioral distinctions preserved by the score and the execution and feedback process in which optimized outputs are used.

\noindent\textbf{Our take-home messages are:}
\begin{enumerate}
\item \textbf{Analyze the scoring chain before making it a target.} Identify which behaviors are omitted, which differences receive the same score, and how execution changes the motion being assessed. Test these potential blind spots with controlled interventions.
\item \textbf{Preserve the behavioral quantities behind the aggregate.} Report continuous measurements alongside thresholded scores, distinguish requested from executed motion, and report progress together with the safety constraints that govern its score.
\item \textbf{Evaluate gain transfer under execution and feedback.} Report how planner outputs are executed. Test whether gains persist when execution changes and planning repeatedly responds to the resulting states, while tracking safety-critical failures.
\item \textbf{Revalidate the complete scoring process.} Changing scoring rules does not by itself address what execution alters or removes. Check whether the revised process captures meaningful behavioral differences and whether optimizing it produces better behavior.
\end{enumerate}

\clearpage
\bibliography{references}
\bibliographystyle{iclr2027_conference}

\appendix
\raggedbottom
\section{Evaluation Protocols and Diagnostics}
\label{app:protocols}

\subsection{Scene Sets and Comparisons}

We use the scene sets defined in Section~\ref{sec:protocols}; the WorldEngine test set follows the official split \citep{worldengine}.
We use the released \texttt{navtest\_failures\_filtered.yaml} scene list at WorldEngine revision \texttt{95002ad}, which contains failure-prone scenarios from \texttt{navtest}.
Open-loop scoring evaluates a four-second plan after LQR tracking with a kinematic bicycle model; surrounding agents replay their logs.
WorldEngine simulation provides observations from the evolving ego state over a four-second rollout with eight updates at 0.5\,s intervals, following four logged history frames; we score the resulting rollout post hoc.
The released entry point uses trajectory playback. The client converts rear-axle poses to world-frame vehicle-center poses and obtains velocities by forward differences on the 10\,Hz trajectory. At each update it applies the pose and motion state at $h=0.5$\,s and returns the resulting state and rendered observations for replanning. In particular, the returned velocity is $[p_c(h+0.1)-p_c(h)]/0.1$, where $p_c$ is the vehicle-center position.
ReCogDrive's 2\,Hz plans are linearly interpolated to this 10\,Hz grid; Qwen-Drive emits native 10\,Hz poses. Both Qwen-Drive checkpoints use reasoning-conditioned inference with one sampled trajectory and seed 42.
The evaluation records closed-loop PDMS, at-fault collisions, and drivable-area exits.
The progress guard is 5.0\,m in open loop and 0.1\,m in closed loop, so absolute EP values are not compared across regimes.
For v1 EP normalization, route progress is first multiplied by the multiplicative penalties. If the maximum of these adjusted progress values across the evaluated proposal and PDM reference exceeds the guard, each is divided by that maximum. Otherwise, EP is one for proposals with a nonzero multiplicative penalty product and zero for those with a zero product.

Open-loop pairs use shared scenes. The primary execution-interface experiment uses 288 scenes in every condition (Appendix~\ref{app:interface-control}). Earlier playback comparisons use 286 jointly completed ReCogDrive scenes, one fewer than the full set for Qwen-Drive, and the full set for NoRD/TOAD; these historical cohorts remain separate.
The nuPlan replication uses \texttt{val14} in both traffic regimes (Section~\ref{sec:nuplan_setup}).

\subsection{Behavioral Measurements}
\label{app:diagnostics}

\paragraph{Sampling and geometry.}
NAVSIM diagnostics use eight predicted positions $p_1,\ldots,p_N$ at $\Delta t=0.5$\,s, with $N=8$.
Smoothing displacement follows Equation~\ref{eq:plan-geometry}, applying one three-point moving average with the first and last predicted positions held fixed.
Its unit is meters; smaller values mean a smaller required positional edit.
The local difference equals one third of the second positional difference and can reflect both bending and uneven speed.
For heading jitter and path length, prepend the ego origin $p_0$ and compute
\begin{equation}
\theta_i=\operatorname{unwrap}\bigl(\operatorname{atan2}(y_i-y_{i-1},x_i-x_{i-1})\bigr),\qquad
L=\sum_{i=1}^{N}\|p_i-p_{i-1}\|_2.
\end{equation}
Heading jitter is the sum in Equation~\ref{eq:plan-geometry}, in radians; it is not a time derivative.
Path length contextualizes headings when displacements are small.

\paragraph{Longitudinal and lateral decomposition.}
The supplementary Qwen decomposition prepends $p_0$ before smoothing, so its first internal point has a different endpoint convention from $D_{\mathrm{smooth}}$.
Let $\delta p_i$ be the edit on this extended sequence and $t_i$ the unit direction of the incoming segment, with perpendicular $n_i$.
The reported components are
\begin{equation}
D_\parallel=\max_i|\delta p_i^\top t_i|,\qquad
D_\perp=\max_i|\delta p_i^\top n_i|.
\end{equation}
The implementation divides each segment by its length plus $10^{-9}$ to handle zero displacement.
These maxima are separate directional readouts, not additive components of the main-table maximum.

\paragraph{Pace and acceleration.}
Planned segment speed and speed change are
\begin{equation}
v_i=\frac{\|p_i-p_{i-1}\|_2}{\Delta t},\qquad
 a_i=\frac{v_{i+1}-v_i}{\Delta t}.
\end{equation}
The Qwen pace table reports signed extrema $\max_i a_i$ and $-\min_i a_i$ as acceleration and deceleration, without clipping them at zero.
ReCogDrive's plan/execution comparison and the retiming experiment use the absolute planned peak $\max_i|a_i|$.
The supplementary planned lateral quantity is $\max_i|y_{i+1}-2y_i+y_{i-1}|/\Delta t^2$ in the initial ego frame.
Longitudinal displacement is $(p(t)-p_0)^\top(\cos\psi_0,\sin\psi_0)$; its time derivative describes longitudinal pace in this fixed frame.
Executed-motion quantities are computed from simulated states separately.

\paragraph{Plan adherence and replanning.}
NAVSIM interpolates the requested trajectory to the simulator's 0.1\,s grid and reports Equation~\ref{eq:adherence} over the four-second horizon, including the initial state.
For nuPlan, let $p^{(t)}(u)$ be the position requested at time $t$ for absolute time $u$.
At each valid planning step $t_i$, compare the plan with the later executed positions at the available horizons $h\in\{0.5,1,\ldots,4\}$\,s:
\begin{equation}
e_{i,h}=\|\hat p(t_i+h)-p^{(t_i)}(t_i+h)\|_2,\quad
D_{\mathrm{adh,max}}=\max_{i,h}e_{i,h},\quad
D_{\mathrm{adh,mean}}=\frac1M\sum_i\max_h e_{i,h}.
\end{equation}
Later nuPlan execution incorporates replanning, so this measurement includes departures induced by subsequent plans.
For plans issued 0.5\,s apart, churn compares matching absolute times:
\begin{equation}
C_i=\frac17\sum_{k=1}^{7}
\|p^{(t_i)}(t_i+0.5+0.5k)-p^{(t_i+0.5)}(t_i+0.5+0.5k)\|_2.
\end{equation}
The scene value is the mean of $C_i$ over valid pairs, with the maximum retained as a supplementary quantity.
Missing plans are excluded, and quantities requiring a planned trajectory are inapplicable to action interfaces without one.
NuPlan plan-geometry scene values average over valid replanning steps.

\paragraph{Executed motion and comfort.}
For an official acceleration or jerk time series $z_r(t)$, an absolute peak is $\max_t|z_r(t)|=\max\{|\min_tz_r(t)|,|\max_tz_r(t)|\}$; this is the operator used for the ReCogDrive comparison and nuPlan jerk results.
Qwen's simulated acceleration and deceleration columns instead report $\max_t a_\parallel(t)$ and $-\min_t a_\parallel(t)$, matching its signed planned extrema.
Equation~\ref{eq:jerk-diagnostic} defines longitudinal jerk; numerical filtering and differentiation follow the official metric implementation.
Comfort checks six channels: longitudinal and lateral acceleration, jerk magnitude, longitudinal jerk, yaw acceleration, and yaw rate.
With channel bounds $\ell_r,u_r$, the NAVSIM binary test is
\begin{equation}
C(\tau)=\prod_{r=1}^{6}\mathbf1[\ell_r<z_r(t)<u_r\ \text{for all evaluated }t].
\end{equation}
Continuous replacements use the bound-utilization ratios $r_k$ specified in Appendix~\ref{app:continuous}.
History comfort applies the same test to the concatenation of logged past states and simulated future states, $\mathrm{HC}=C(\tau_{\mathrm{past}}\oplus\tau_{\mathrm{exec}})$.
For two successive simulated predictions aligned at their overlapping absolute times, let $z^{(1)}_r(t_i)$ and $z^{(2)}_r(t_i)$ be acceleration magnitude, jerk magnitude, yaw rate, or yaw acceleration.
The official extended-comfort term is
\begin{equation}
\mathrm{EC}=\prod_r\mathbf1\left[\sqrt{\frac1n\sum_{i=1}^{n}\bigl(z^{(1)}_r(t_i)-z^{(2)}_r(t_i)\bigr)^2}\le b_r\right].
\end{equation}
The four bounds are $0.7$\,m/s$^2$, $0.5$\,m/s$^3$, $0.1$\,rad/s, and $0.1$\,rad/s$^2$, respectively.
The two-stage experiment uses this term for the aligned prediction pairs in the official protocol (Appendix~\ref{app:pairs-twostage}).

\paragraph{Driving direction.}
Let $I_i^{\mathrm{opp}}$ be the scorer's oncoming-traffic mask.
The largest accumulated displacement in its one-second sliding window is
\begin{equation}
d_{\mathrm{opp}}=\max_j\sum_{i=\max(0,j-H)}^j\|p_i-p_{i-1}\|_2 I_i^{\mathrm{opp}},
\end{equation}
where the initial increment is zero and $H=1\,\mathrm{s}/\Delta t_{\mathrm{sim}}$.
The score is 1 for $d_{\mathrm{opp}}<2$\,m, 0.5 for $2\le d_{\mathrm{opp}}<6$\,m, and 0 otherwise.
The v2 implementation removes intersection positions from the mask; the v1 and v2 DDC values are therefore reported separately.
Map-verified wrong-way distance in the controlled probe is a separate measurement from this thresholded score.

\paragraph{Lane keeping.}
Let $d_i=\operatorname{dist}(p_i,\mathcal C)$ for the route centerline $\mathcal C$.
Starting from $q_0=0$, the implementation updates the exceedance counter as
\begin{equation}
q_i=\begin{cases}
q_{i-1},&p_i\text{ is in an intersection},\\
q_{i-1}+1,&d_i>0.5\,\mathrm m\text{ outside intersections},\\
0,&\text{otherwise},
\end{cases}\qquad
\mathrm{LK}=\mathbf1\!\left[\max_iq_i<\left\lceil\frac{2\,\mathrm s}{\Delta t_{\mathrm{sim}}}\right\rceil\right].
\end{equation}
This differs from drivable-area compliance, which checks the vehicle footprint against the drivable region.

\paragraph{Outcome rates and reported progress.}
For outcome $f$ (at-fault collision or road departure), let $I_s^f$ indicate its occurrence in scene $s$.
Over $n$ matched scenes, the count is $N_f=\sum_{s=1}^n I_s^f$ and the rate is $r_f=N_f/n$.
They are distinct from mean penalty values, since NC includes a partial penalty for static-object collisions.
Equation~\ref{eq:ep-decomposition} defines the reported-progress decomposition; comparisons conditional on both arms passing use their intersection, rather than each arm's own passing set.
All paired differences are optimized minus baseline in the metric's native units.

\subsection{Descriptive Planner Profiles}
\label{app:teaser-profiles}

Figure~\ref{fig:teaser}(b) groups 24 configurations into IL (13), RL (5), score distillation (5), and CEM (1): 22 checkpoints and the TOAD checkpoint evaluated without and with search.
NoRD-SFT belongs to IL; the Qwen-Drive pair is reported separately in Table~\ref{tab:pairs}.
Configurations using multiple procedures are assigned once with priority CEM, RL, distillation, then IL.
Each profile averages configuration-level ratios with equal weight; planner composition differs across groups, so the profiles are descriptive.

For scored components, lane keeping, and direction, $q$ is the planner mean divided by the logged-human mean on the same scenes.
For smoothing displacement and plan-adherence error, the ratio is reversed.
Human $=1$ denotes equality with the logged-human mean on the same scenes.
All axes map ratios $(0,0.5,0.8,1.1,1.27)$ to radii $(0,0.12,0.28,0.88,1)$ by piecewise-linear interpolation.
The upper half contains components that influence PDMS; the lower half contains zero-weight DDC, single-stage NAVSIM-v2 lane keeping, smoothing displacement, and plan adherence.

\subsection{Paired Statistical Analysis}
\label{app:uncertainty}

Mean changes use paired scene-bootstrap 95\% intervals with 2{,}000 resamples.
Log-bootstrap intervals resample driving logs, retaining their scenes together to account for within-log dependence.
Wilcoxon signed-rank tests use Holm correction over the original analysis families: 42 geometry/score tests and 30 EPDMS tests for the six training comparisons, and 24 tests across TOAD's augmentation and search arms.
The displayed tables select quantities relevant to the main text; these selections do not change the correction families.
A small signed-rank $p$-value need not imply a nonzero mean change, so confidence intervals guide interpretation of mean effects.
Qwen-Drive's DDC change, NoRD's plan-adherence change, GTRS-Dense's jitter change, and TOAD's \texttt{navtest} PDMS change have log intervals including zero.
We retain these cases alongside supported improvements and declines.

In Table~\ref{tab:pairs}, $^{\dagger}$ marks changes for which either interval includes zero, the interval directions disagree, or the Holm-adjusted test does not pass $p<0.05$. For nuPlan, support likewise requires the scene and log intervals to exclude zero in the same direction and the Wilcoxon test to pass Holm correction over the original 38-quantity family in each traffic regime.
The family is retained even for measurements omitted from the final tables.
These intervals quantify evaluation-scene and driving-log variability for the fixed released checkpoints.

\subsection{Planner-Specific Settings}
\label{app:planner-settings}

\paragraph{NoRD.}
NoRD uses a fixed human-derived trajectory vocabulary and PDMS rewards during reinforcement fine-tuning \citep{nord2026}.
This specifies the action representation for the comparison; the observed diagnostic changes are reported in Table~\ref{tab:pairs}.

\paragraph{Qwen-Drive.}
\label{app:qwen-protocol}
The reinforcement objective uses EP/TTC/comfort weights $6/4/2$ and a displacement reward toward the logged trajectory, with exploration in six low-frequency cosine modes \citep{qwendrive2026}.
Both released experts use reasoning-conditioned inference, matching the reinforcement checkpoint's training mode; direct SFT inference supplies the conditioning control in Appendix~\ref{app:qwen}.
The expert emits 50 poses at 10\,Hz, from which eight poses at 0.5\,s intervals are selected for scoring.
Inputs use the three released camera views at the four most recent 2\,Hz frames and 10\,Hz ego history from the nuPlan database, filtered according to the released model documentation.
Scoring the leading 40 poses instead changes PDMS by 0.03 points.
For WorldEngine replay, both experts use the same four rendered camera frames, with the required 10\,Hz ego history interpolated from the available 2\,Hz states; the leading 40 predicted poses are passed to the standard trajectory interface.
Repeated deterministic inference reproduces the trajectories bit-identically.
Reproduced PDMS is 87.7, 87.9, and 90.5 for direct SFT, reasoning-conditioned SFT, and RL, respectively, against published values of 87.8, 88.2, and 90.7; all sub-metrics agree within 0.5 points \citep{qwendrive2026}.

\paragraph{TOAD on NAVSIM-v1.}
\label{app:toad-protocol}
The released DrivoR checkpoint uses 64 planner proposals, 64 random control-space perturbations, and CEM with 64 samples, 8 elites, and 10 rounds.
Perturbations are divided equally among constant, linearly varying, and free-form controls.
Search combines predicted sub-scores through the PDMS rule, with zero DDC weight, a comfort penalty of weight 0.05, and an anchoring penalty.
The CEM mean is deployed only when its predicted score exceeds that of the best proposal.
We seed perturbations per scene from its token, making the baseline, augmentation-only, and search arms independent of processing order.
All recorded trajectories are evaluated by the same official scorers.
The v2 direction comparison and search configuration are specified in Appendix~\ref{app:toad-ddc-comparison}.

\paragraph{Plan-R1 on nuPlan.}
\label{app:nuplan-protocol}
\label{app:nuplan-planr1}
Evaluation uses 15\,s scenarios, 10\,Hz replanning, LQR execution, and logged or IDM-reactive traffic.
The first plan is also evaluated without replanning using the CLS-style PDM score.
Both checkpoints share a 1{,}024-token vocabulary and a frozen prediction branch; all 482 tensors of that branch match between the released files.
Fine-tuning samples $G=4$ sequences and uses
\begin{equation}
r_t=g^{\mathrm{road}}_t g^{\mathrm{agent}}_t g^{\mathrm{obstacle}}_t
\frac{2c_t+5\tau_t+2s+p}{10},
\label{eq:planr1-reward}
\end{equation}
where the gates penalize road departures and collisions, $c_t$ and $\tau_t$ are comfort and TTC, and $s$ and $p$ are rollout speed-limit compliance and progress.
Group-centered rewards-to-go use fixed scaling 0.1; the KL penalty toward the frozen branch also has weight 0.1.
The reward changes CLS weights and omits its making-progress and direction gates; comfort remains thresholded, without a continuous jerk penalty.
Executed peaks use $\max_t|x_t|$, computed from both extrema of each time series.
Plan geometry uses each plan's first four seconds; churn compares plans issued 0.5\,s apart at matching absolute times.

\section{Metric Sensitivity}
\label{app:blindspots}

\subsection{Comfort and Tracker Effects}
\label{app:comfort-tracker}

\begin{figure}[H]
\centering
\includegraphics[width=\linewidth]{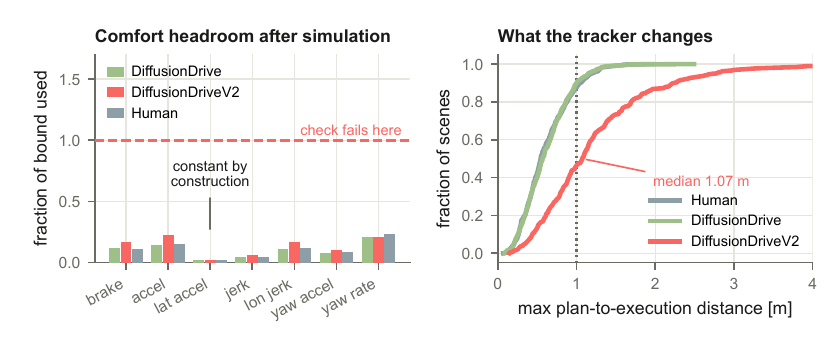}
\caption{
\textbf{Comfort sensitivity and plan adherence on \texttt{navhard}.}
\textbf{Left:} Mean utilization of each comfort bound for DiffusionDrive v1, v2, and logged Human trajectories; a value of one denotes the bound. The lateral-acceleration check is constant by construction.
\textbf{Right:} Empirical cumulative distribution of the maximum frame-wise distance between planned waypoints and LQR-executed positions, in meters.
}
\label{fig:absorbed}
\end{figure}

Binary comfort fails on 410 of 291{,}504 \texttt{navtest} planner--scene evaluations (0.14\%).
One of its six checks is constant because the motion model's lateral-acceleration channel is zero.
For DiffusionDrive, DiffusionDriveV2, and the human trajectory on \texttt{navhard}, the largest within-scene comfort utilization averages 0.32--0.38 of its threshold.
Figure~\ref{fig:absorbed} relates this margin to the difference between planned and tracked trajectories.
For ReCogDrive-2B, the mean absolute planned acceleration peak rises from 1.14 to 2.84\,m/s$^2$ after fine-tuning, while the absolute executed longitudinal peak rises from 0.88 to 1.20\,m/s$^2$.
The corresponding 8B values are 1.14 to 2.95 planned and 0.89 to 1.34 executed.

\subsection{Continuous Comfort Rescoring}
\label{app:continuous}

For continuous rescoring, the two-sided longitudinal-acceleration check is separated into braking and propulsion, giving seven bound utilizations $r_k$.
For an upper bound $b_k>0$, $r_k=\max(0,\max_t z_k(t)/b_k)$; for the lower braking bound $-b_k$, $r_k=\max(0,-\min_t a_\parallel(t)/b_k)$; magnitude checks use $r_k=\max_t|z_k(t)|/b_k$.
The bounds for braking, propulsion, lateral acceleration, jerk magnitude, longitudinal jerk, yaw acceleration, and yaw rate are $4.05$, $2.40$, $4.89$, $8.37$, $4.13$, $1.93$, and $0.95$, in their corresponding SI units.
Using the same fixed trajectories, we replace binary comfort by three functions of these utilizations:
\begin{align}
C_{\rm worst}&=\mathrm{clip}(1-\max_k r_k,0,1),\\
C_{\rm sigmoid}&=\sigma\bigl((1-\max_k r_k)/0.1\bigr),\\
C_{\rm mean}&=K^{-1}\sum_k\mathrm{clip}(1-r_k,0,1).
\end{align}
Here $K=7$ and $\sigma(x)=1/(1+e^{-x})$; all other PDMS components are unchanged.
The ranking of DiffusionDrive, DiffusionDriveV2, and Human remains unchanged (Table~\ref{tab:continuous}); the worst-channel margin retains 65\% and 75\% of the original DiffusionDriveV2 advantage on the two sets.
This fixed-output comparison measures how each comfort definition separates the three planners.

\begin{table}[H]
\centering
\caption{
\textbf{Comfort rescoring on \texttt{navhard} and the WorldEngine test set.} Mean PDMS ($\uparrow$) under the official and continuous comfort formulations, using fixed outputs.
DD denotes DiffusionDrive and DDv2 denotes DiffusionDriveV2.
Retained reports the DDv2--DD difference relative to the official binary difference, computed from unrounded means.
}
\label{tab:continuous}
\tablelayout
\begin{tabular*}{\linewidth}{@{\extracolsep{\fill}}lrrrrrrrr@{}}
\toprule
& \multicolumn{4}{c}{\texttt{navhard}}
& \multicolumn{4}{c}{WorldEngine test set} \\
\cmidrule(lr){2-5}
\cmidrule(lr){6-9}
Comfort term
& DD & DDv2 & Human & Retained
& DD & DDv2 & Human & Retained \\
\midrule
Official (binary)
& 0.7044 & 0.7328 & 0.8992 & 100\%
& 0.6540 & 0.6960 & 0.9325 & 100\% \\
Worst-channel margin
& 0.6616 & 0.6801 & 0.8458 & 65\%
& 0.6144 & 0.6458 & 0.8778 & 75\% \\
Smooth sigmoid
& 0.7038 & 0.7315 & 0.8982 & 97\%
& 0.6534 & 0.6948 & 0.9313 & 98\% \\
Mean margin
& 0.6903 & 0.7138 & 0.8814 & 83\%
& 0.6410 & 0.6777 & 0.9143 & 87\% \\
\bottomrule
\end{tabular*}
\end{table}

\subsection{Driving-Direction Construction}
\label{app:wrongway}

A route-centerline follower is compared with offsets of $\{\pm1.75,\pm3.5\}$\,m at the same speed profile.
Matched trajectories pass NC and DAC, differ in raw progress by less than 0.1\,m, and have identical TTC and comfort.
Lane direction at flagged frames is verified against map lane baselines, accounting for the different traffic conventions in Singapore and the US maps.
The construction yields 42 pairs across 37 \texttt{navhard} scenes, with 8.80\,m of mean verified wrong-way travel; DDC is zero for 31 of the 42 violating trajectories.
Mean paired PDMS differs by $+0.000019$ and the median difference is zero.
Re-aggregating the same sub-scores with DDC weight 2 or 5 gives differences of $-0.157$ or $-0.278$; using DDC as a gate gives $-0.797$.

\subsection{Positional Smoothing and Re-timing}
\label{app:smoothdecomp}
\label{app:smoothing-results}

Positional smoothing applies one three-point moving average to waypoints with fixed endpoints and preserves their original headings.
Holding headings fixed isolates the positional edit from heading reconstruction.
For DiffusionDriveV2 it removes 70\% of heading jitter on \texttt{navhard} and 74\% on the WorldEngine test set, with mean waypoint displacements of 0.417 and 0.409\,m.
PDMS changes by $-0.003$ $[-0.013,+0.007]$ and $+0.007$ $[-0.006,+0.021]$, respectively.
The same edit changes DiffusionDrive PDMS by $+0.004$ and $-0.002$, and Human PDMS by $-0.005$ and $-0.016$ on the respective sets.

\begin{figure}[!t]
\centering
\includegraphics[width=\linewidth]{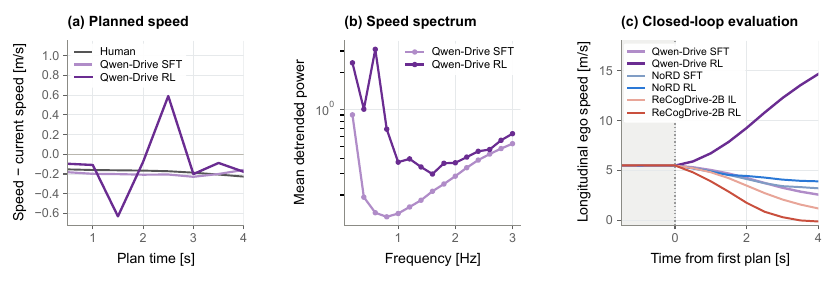}
\caption{
\textbf{Planned and executed speed.} (a) Mean planned speed relative to ego speed on \texttt{navtest}. (b) Mean speed spectrum after linear detrending (5\,s, 10\,Hz). (c) Mean longitudinal speed in WorldEngine NR simulation \citep{worldengine}; matched sets: Appendix~\ref{app:protocols}. Shading: logged history; zero: first action.
}
\label{fig:qwenpace}
\end{figure}

\paragraph{Qwen-Drive controls.}
\label{app:qwen}
Switching SFT from direct to reasoning-conditioned inference changes PDMS by $+0.002$ $[+0.001,+0.004]$, smoothing displacement by $+0.000$\,m, and plan-adherence error by $-0.001$\,m.
This conditioning comparison does not account for the smoothing-displacement increase between the SFT and RL checkpoints in Table~\ref{tab:qwen-pace}.
The larger longitudinal component locates most of the variation along the path.

\begin{table}[H]
\centering
\caption{
\textbf{NAVSIM-v1 \texttt{navtest}: plan and simulated motion.}
Geometry and pace of the Qwen-Drive experts and the human trajectory: smoothing displacement and its components along and across the local path direction, planned peak acceleration and deceleration between consecutive 0.5\,s segments, and the peak acceleration and deceleration of the simulated states that the scorer evaluates.
}
\label{tab:qwen-pace}
\tablelayout
\begin{tabular*}{\linewidth}{@{\extracolsep{\fill}}lrrrrrrr@{}}
\toprule
 & \multicolumn{3}{c}{Smoothing displacement [m] $\downarrow$} & \multicolumn{2}{c}{Planned [m/s$^2$]} & \multicolumn{2}{c}{Simulated [m/s$^2$]} \\
\cmidrule(lr){2-4}\cmidrule(lr){5-6}\cmidrule(lr){7-8}
Checkpoint & Total & Along & Across & Acc. & Dec. & Acc. & Dec. \\
\midrule
Qwen-Drive SFT & $0.116$ & $0.103$ & $0.049$ & $0.67$ & $0.66$ & $0.32$ & $0.41$ \\
Qwen-Drive RL & $0.212$ & $0.204$ & $0.059$ & $1.65$ & $1.88$ & $0.28$ & $0.53$ \\
Human & $0.099$ & $0.081$ & $0.050$ & $0.37$ & $0.43$ & $0.32$ & $0.43$ \\
\bottomrule
\end{tabular*}
\end{table}

Re-timing smooths segment speeds along each trajectory's own path, preserving endpoint speeds and total path length.
On \texttt{navtest} it reduces the RL checkpoint's largest planned speed change per unit time from 2.46 to 1.31\,m/s$^2$, with mean PDMS change $-0.0002$ and an interval including zero (Table~\ref{tab:qwen-retime}).
The SFT and Human controls also have small mean score changes under this intervention.

\begin{table}[H]
\centering
\caption{
\textbf{NAVSIM-v1 \texttt{navtest}: retiming intervention.}
Re-timing the deployed trajectory along its own path: largest speed change per unit time between consecutive segments before and after, PDMS before and after, and the paired change with its bootstrap 95\% interval.
}
\label{tab:qwen-retime}
\tablelayout
\begin{tabular*}{\linewidth}{@{\extracolsep{\fill}}lrrrrr@{}}
\toprule
 & \multicolumn{2}{c}{$\max|\Delta v|/\Delta t$ [m/s$^2$] $\downarrow$} & \multicolumn{3}{c}{PDMS $\uparrow$} \\
\cmidrule(lr){2-3}\cmidrule(lr){4-6}
Checkpoint & Original & Retimed & Original & Retimed & $\Delta$ [95\% CI] \\
\midrule
Qwen-Drive SFT & $1.24$ & $0.91$ & $0.8791$ & $0.8790$ & $-.0002$ \ci{-.0005}{.0002} \\
Qwen-Drive RL & $2.46$ & $1.31$ & $0.9047$ & $0.9045$ & $-.0002$ \ci{-.0008}{.0003} \\
Human & $0.97$ & $0.87$ & $0.9455$ & $0.9453$ & $-.0002$ \ci{-.0005}{.0000} \\
\bottomrule
\end{tabular*}
\end{table}

\subsection{Progress and Candidate Ties}
\label{app:progress-decomp}
\label{app:nord-pace}
\label{app:qwen-pace}
\label{app:saturation}

Equation~\ref{eq:ep-decomposition} separates reported EP into pass frequency and conditional progress.
NoRD's EP increases by 0.091 as its pass rate rises from 81.7\% to 92.3\%, while conditional EP changes from 0.846 to 0.847.
Its mean planned speed falls by 0.16\,m/s and path length by 0.66\,m.
Qwen-Drive instead gains 0.027 EP with both pass rate (94.9\% to 96.8\%) and conditional EP (0.865 to 0.876) increasing; mean speed rises by 0.10\,m/s and path length by 0.40\,m.
These comparisons distinguish a change in pass frequency from a change in progress among passed scenes.

In the fixed 800-candidate DiffusionDriveV2 pool on \texttt{navhard}, 274 scenes have tied PDMS maxima and 269 have tied EPDMS maxima.
Mean tied-set sizes are 146.6 and approximately 137.
Within the respective tied sets, heading jitter spans 0.43--7.29 and 0.45--5.63, while peak lateral acceleration spans 0.98--6.29 and 0.99--5.61\,m/s$^2$.
Both scores therefore leave geometric differences unresolved among some maximizers.

\section{Paired Optimization Responses}
\label{app:geometry}

\subsection{Scene-Set Replications}
\label{app:paired-replications}

Table~\ref{tab:pairs-replicated} combines the \texttt{navhard} and WorldEngine test-set replications, reporting PDMS changes alongside fixed-output EPDMS rescoring and behavioral diagnostics.
ReCogDrive's geometric changes and NoRD's reductions in jitter and smoothing displacement recur on both sets.
Qwen-Drive again increases smoothing displacement while reducing plan-adherence error; the GTRS-Dense changes remain small on these quantities.

\begin{table}[H]
\caption{\textbf{Paired changes on safety-critical cases.} Scores, EPDMS components, and behavioral diagnostics on fixed planner outputs from \texttt{navhard} and the WorldEngine test set. PDMS uses the v1 scorer; EPDMS and its components use the v2 scorer. DDC$^{\mathrm{v1}}$ has zero PDMS weight, whereas DDC$^{\mathrm{v2}}$ gates EPDMS and excludes intersection positions. Jitter is in radians; smoothing displacement (Smooth.) and plan-adherence error (Adh.) are in meters. Gray entries have unadjusted paired-test $p\geq0.05$ in these subset analyses; \texttt{navtest} adjusted tests and intervals appear in Tables~\ref{tab:pairs-uncertainty} and~\ref{tab:v2pairs-uncertainty}.}
\label{tab:pairs-replicated}
\centering\tablelayout
\setlength{\tabcolsep}{2pt}
\begin{tabular*}{\linewidth}{@{\extracolsep{\fill}}lrrrrrrrrr@{}}
\toprule
& \multicolumn{2}{c}{Aggregate scores} & \multicolumn{3}{c}{EPDMS components} & \multicolumn{4}{c}{Behavioral diagnostics} \\
\cmidrule(lr){2-3}\cmidrule(lr){4-6}\cmidrule(lr){7-10}
Pair & \metrichead{PDMS}{\uparrow} & \metrichead{EPDMS}{\uparrow} & \metrichead{DDC$^{\mathrm{v2}}$}{\uparrow} & \metrichead{LK}{\uparrow} & \metrichead{HC}{\uparrow} & \metrichead{DDC$^{\mathrm{v1}}$}{\uparrow} & \metrichead{Jitter}{\downarrow} & \metrichead{Smooth.}{\downarrow} & \metrichead{Adh.}{\downarrow} \\

\midrule
\multicolumn{10}{l}{\textit{\texttt{navhard}}} \\
ReCog-2B IL$\to$RL & $+.105$ & $+.077$ & $-.029$ & $-.044$ & \ns{$-.004$} & $-.032$ & $+0.83$ & $+.154$ & $+.385$ \\
ReCog-8B IL$\to$RL & $+.079$ & $+.060$ & $-.016$ & \ns{$-.024$} & \ns{$-.004$} & $-.012$ & $+0.71$ & $+.164$ & $+.399$ \\
Qwen-Drive SFT$\to$RL & $+.063$ & $+.062$ & \ns{$-.001$} & \ns{$+.016$} & \ns{$+.000$} & \ns{$-.001$} & $+.03$ & $+.091$ & $-.067$ \\
DDv1$\to$v2 & $+.028$ & \ns{$-.000$} & $-.017$ & \ns{$-.011$} & $-.096$ & $-.024$ & $+1.25$ & $+.310$ & $+.609$ \\
NoRD IL$\to$RL & $+.152$ & $+.145$ & $+.050$ & $+.056$ & $+.031$ & $+.046$ & $-0.29$ & $-.052$ & \ns{$-.053$} \\
\shortstack[l]{GTRS-Dense\\expert$\to$reward} & \ns{$+.004$} & $-.003$ & \ns{$-.011$} & \ns{$-.022$} & \ns{$+.007$} & \ns{$-.014$} & \ns{$-.05$} & \ns{$-.001$} & $-.089$ \\
\midrule
\multicolumn{10}{l}{\textit{WorldEngine test set}} \\
ReCog-2B IL$\to$RL & $+.093$ & $+.079$ & $-.028$ & \ns{$-.028$} & \ns{$.000$} & $-.035$ & $+1.07$ & $+.156$ & $+.358$ \\
ReCog-8B IL$\to$RL & $+.057$ & $+.034$ & \ns{$-.009$} & \ns{$-.014$} & \ns{$-.003$} & \ns{$-.010$} & $+0.84$ & $+.167$ & $+.412$ \\
Qwen-Drive SFT$\to$RL & $+.050$ & $+.043$ & \ns{$+.000$} & \ns{$+.007$} & \ns{$+.003$} & \ns{$-.003$} & $+.07$ & $+.092$ & $-.066$ \\
DDv1$\to$v2 & $+.042$ & \ns{$+.014$} & $-.028$ & \ns{$-.003$} & $-.094$ & $-.030$ & $+1.46$ & $+.300$ & $+.581$ \\
NoRD IL$\to$RL & $+.159$ & $+.139$ & $+.061$ & $+.087$ & $+.045$ & $+.045$ & $-0.34$ & $-.064$ & \ns{$-.074$} \\
\shortstack[l]{GTRS-Dense\\expert$\to$reward} & \ns{$+.013$} & \ns{$-.007$} & \ns{$-.009$} & $-.024$ & \ns{$+.014$} & $-.021$ & \ns{$+.05$} & \ns{$-.004$} & $-.106$ \\
\bottomrule
\end{tabular*}
\end{table}

\subsection{Log-Level Uncertainty}

Tables~\ref{tab:pairs-uncertainty} and~\ref{tab:v2pairs-uncertainty} provide the uncertainty estimates for the paired changes in Table~\ref{tab:pairs}.
TOAD's unsearched and searched arms use the same weights; the original correction family also includes the augmentation-only arm, whose PDMS change is approximately zero.
The search arm's log interval includes zero for PDMS, while its geometric changes have mixed directions.
The EPDMS intervals for the DiffusionDrive IL-to-RL transition and GTRS-Dense also include zero.

\begin{table}[H]
\caption{
\textbf{NAVSIM-v1 \texttt{navtest}: paired uncertainty.} Holm-adjusted Wilcoxon $p$-values (first row), scene-bootstrap 95\% intervals (second row), and driving-log-bootstrap 95\% intervals (third row) for the five v1 quantities of Table~\ref{tab:pairs}. Correction uses the original families specified in Appendix~\ref{app:uncertainty}, including TOAD's augmentation-only arm.
}
\label{tab:pairs-uncertainty}
\centering
\tablelayout
\begin{tabular*}{\linewidth}{@{\extracolsep{\fill}}lrrrrr@{}}
\toprule
Pair & PDMS & DDC & Jitter & Smooth. & Adh. \\
\midrule
ReCog-2B IL$\to$RL
 & $<\!10^{-4}$ & $<\!10^{-4}$ & $<\!10^{-4}$ & $<\!10^{-4}$ & $<\!10^{-4}$ \\
{\scriptsize Scene CI} & \ci{.038}{.046} & \ci{-.015}{-.011} & \ci{1.15}{1.23} & \ci{.180}{.185} & \ci{.311}{.326} \\
{\scriptsize Log CI} & \ci{.035}{.050} & \ci{-.021}{-.008} & \ci{1.07}{1.31} & \ci{.176}{.189} & \ci{.287}{.351} \\
ReCog-8B IL$\to$RL
 & $<\!10^{-4}$ & $<\!10^{-4}$ & $<\!10^{-4}$ & $<\!10^{-4}$ & $<\!10^{-4}$ \\
{\scriptsize Scene CI} & \ci{.031}{.038} & \ci{-.007}{-.004} & \ci{.88}{.95} & \ci{.171}{.174} & \ci{.391}{.406} \\
{\scriptsize Log CI} & \ci{.029}{.040} & \ci{-.008}{-.003} & \ci{.84}{1.00} & \ci{.168}{.177} & \ci{.372}{.427} \\
Qwen-Drive SFT$\to$RL
 & $<\!10^{-4}$ & $0.013$ & $<\!10^{-4}$ & $<\!10^{-4}$ & $<\!10^{-4}$ \\
{\scriptsize Scene CI} & \ci{.023}{.029} & \ci{-.002}{-.000} & \ci{.05}{.09} & \ci{.095}{.098} & \ci{-.048}{-.043} \\
{\scriptsize Log CI} & \ci{.021}{.030} & \ci{-.003}{.000} & \ci{.04}{.10} & \ci{.093}{.101} & \ci{-.053}{-.038} \\
DDv1$\to$v2
 & $<\!10^{-4}$ & $<\!10^{-4}$ & $<\!10^{-4}$ & $<\!10^{-4}$ & $<\!10^{-4}$ \\
{\scriptsize Scene CI} & \ci{.025}{.032} & \ci{-.015}{-.011} & \ci{1.32}{1.41} & \ci{.277}{.291} & \ci{.641}{.670} \\
{\scriptsize Log CI} & \ci{.023}{.034} & \ci{-.019}{-.007} & \ci{1.29}{1.45} & \ci{.270}{.298} & \ci{.620}{.694} \\
\midrule
NoRD IL$\to$RL
 & $<\!10^{-4}$ & $<\!10^{-4}$ & $<\!10^{-4}$ & $<\!10^{-4}$ & $<\!10^{-4}$ \\
{\scriptsize Scene CI} & \ci{.101}{.114} & \ci{.013}{.020} & \ci{-.24}{-.19} & \ci{-.034}{-.027} & \ci{-.007}{.013} \\
{\scriptsize Log CI} & \ci{.096}{.120} & \ci{.012}{.021} & \ci{-.26}{-.18} & \ci{-.036}{-.025} & \ci{-.013}{.019} \\
\midrule
GTRS-Dense expert$\to$reward
 & $<\!10^{-4}$ & $<\!10^{-4}$ & $0.013$ & $<\!10^{-4}$ & $<\!10^{-4}$ \\
{\scriptsize Scene CI} & \ci{.007}{.014} & \ci{-.006}{-.003} & \ci{-.00}{.04} & \ci{-.002}{-.001} & \ci{-.065}{-.035} \\
{\scriptsize Log CI} & \ci{.005}{.015} & \ci{-.006}{-.002} & \ci{-.02}{.06} & \ci{-.002}{-.000} & \ci{-.080}{-.018} \\
\midrule
TOAD & $0.79$ & $<\!10^{-4}$ & $0.0053$ & $<\!10^{-4}$ & $0.00016$ \\
{\scriptsize Scene CI} & \ci{.000}{.002} & \ci{-.005}{-.003} & \ci{-.16}{-.12} & \ci{.002}{.003} & \ci{.005}{.010} \\
{\scriptsize Log CI} & \ci{-.000}{.002} & \ci{-.006}{-.003} & \ci{-.17}{-.11} & \ci{.002}{.003} & \ci{.004}{.011} \\
\bottomrule
\end{tabular*}
\end{table}

\subsection{Qualitative Cases and Search Response}
\label{app:geometry-qualitative}
\label{app:toad-sweep}

\begin{figure}[H]
\centering
\includegraphics[width=\linewidth]{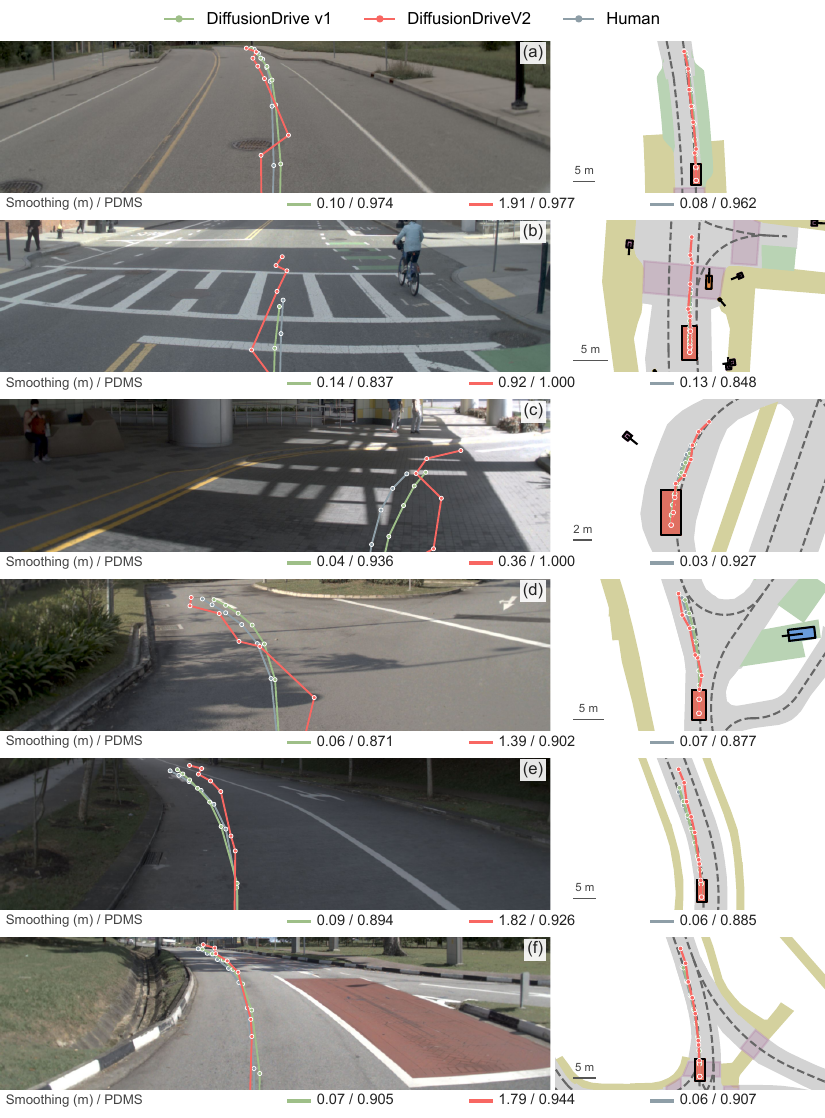}
\caption{
\textbf{Additional illustrative trajectory comparisons.} Six selected WorldEngine test scenes \citep{worldengine}. Each row pairs a cropped front-camera view with a bird's-eye view; curves show planned waypoints. Values give smoothing displacement (m) / PDMS in legend order. In every scene, v2 scores above both v1 and Human while all three pass NC, DAC, TTC, and comfort after tracking. BEV scale bars indicate the per-scene zoom. These cases illustrate the score--geometry mismatch; Table~\ref{tab:pairs} reports aggregate diagnostic changes.
}
\label{fig:qualitative}
\end{figure}

Figure~\ref{fig:qualitative} adds six scenes beyond Figure~\ref{fig:qualitative-selected}, showing oscillatory DiffusionDriveV2 plans that score above both v1 and Human.

\begin{figure}[H]
\centering
\includegraphics[width=\linewidth]{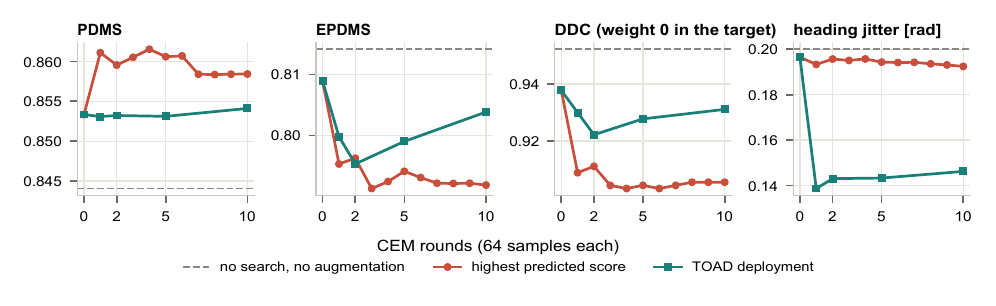}
\caption{
\textbf{TOAD search-depth sweep on \texttt{navhard}.}
Fixed DrivoR weights and 64 samples per CEM round are used throughout.
Panels report official PDMS, EPDMS, DDC, and heading jitter (radians).
Squares denote TOAD's deployed trajectory; circles denote the candidate with the highest predicted PDMS among all evaluated candidates, starting from the augmented proposal set at round zero.
Dashed lines show the planner's argmax without augmentation or search.
Candidate ranking uses the learned score; the plotted scores are obtained by evaluating the selected trajectories.
}
\label{fig:toadsweep}
\end{figure}

Figure~\ref{fig:toadsweep} separates TOAD's deployed trajectory from the candidate with the highest predicted score.
On \texttt{navhard}, that candidate gains 0.008 official PDMS within one CEM round while DDC falls from 0.94 to 0.91; subsequent rounds produce little additional change.
For the same candidate sequence, EPDMS falls from $0.809$ at round zero to $0.795$ after one round.
The \texttt{navtest} deployed-arm comparison is reported in Tables~\ref{tab:pairs} and~\ref{tab:v2pairs-uncertainty}.

At ten rounds, DDC is 0.931 for the deployed arm and 0.906 for the highest-predicted-score candidate in the sequential trace.
TOAD deploys the converged CEM mean subject to its acceptance rule, with comfort and anchoring penalties in the search objective; the comparison evaluates that complete deployment procedure.

\paragraph{Direction under the v2 search objective.}
\label{app:toad-ddc-comparison}
We compare the same released TOAD v2 checkpoint without search and with proposal augmentation and CEM (64 samples per round, 8 elites, 5 rounds) \citep{xu2026toad}.
The predicted-component weights are 10 (NC), 13 (DAC), 6 (DDC), 14 (TTC), 15 (EP), and 2 (comfort).
We use the official NAVSIM-v2 \texttt{navhard-two-stage} protocol with 450 original scenes and 5{,}462 synthetic scenes initialized at counterfactual ego states \citep{cao2025pseudo}.
Mean deployed DDC changes from 0.9933 to 0.9944 on the original scenes and from 0.9369 to 0.9388 on the synthetic scenes.
DDC is included in this target, and its mean does not decline under search in either stage.
The v1/v2 comparison evaluates the two released configurations, whose checkpoints and other objective weights also differ.

\section{Closed-Loop Evaluation and Cross-Benchmark Validation}
\label{app:closedloop}
\label{app:nuplan}

\subsection{Full Matched Execution-Interface Comparison}
\label{app:interface-control}
Table~\ref{tab:execution-interface} evaluates all four conditions on the common scene set per family (1,152 scores), retaining failures. The only experimental switch within each family is the execution interface; checkpoints, scene selection, inference settings, horizon, traffic, and scoring are held fixed. The LQR arm uses NAVSIM's tracker and bicycle motion model, returning the resulting motion state for replanning. Both arms use the same post-hoc scoring configuration. There are 576 identical initial-plan pairs per family, and saved plans, returned frames, and LQR pose consistency were checked. Seeds are fixed; the intervals below describe scene/log variability for released checkpoints, not training or seed variability.

Qwen-Drive and TOAD were run locally; ReCogDrive-2B/8B and NoRD were run on the remote host. Each four-condition comparison is entirely within one host. The two ReCogDrive families use the existing unknown-navigation-command-to-straight adapter in all arms, including the affected scenes. NoRD's full comparison disables prefix caching to restore repeatable initial predictions; the failed attempt was retained and the comparison rerun. These full NR results are distinct from the earlier playback cohorts below, and are not combined with their open-loop or reactive results to form a new paired contrast.

\begin{table}[!ht]
\centering\tablelayout
\caption{\textbf{Paired PDMS gains within each execution interface.} Mean [log-bootstrap 95\% CI], the full scene set per row. Interaction estimates and intervals appear in the main comparison (Table~\ref{tab:execution-interface}). Each bootstrap resamples driving logs with all their scenes (2,000 draws, seed 0). Intervals are pointwise.}
\label{tab:interface-intervals}
\begin{tabular*}{\linewidth}{@{\extracolsep{\fill}}lrr@{}}
\toprule
Planner & $\Delta S_{\mathrm{PB}}$ & $\Delta S_{\mathrm{LQR}}$ \\
\midrule
ReCogDrive-2B & $-0.236$ [-0.300, -0.169] & $+0.218$ [+0.165, +0.273] \\
ReCogDrive-8B & $-0.200$ [-0.265, -0.132] & $+0.158$ [+0.101, +0.214] \\
Qwen-Drive & $-0.388$ [-0.457, -0.321] & $+0.136$ [+0.071, +0.198] \\
NoRD & $+0.156$ [+0.102, +0.210] & $+0.186$ [+0.128, +0.244] \\
TOAD & $+0.093$ [+0.043, +0.137] & $+0.073$ [+0.042, +0.104] \\
\bottomrule
\end{tabular*}
\end{table}

\begin{table}[!ht]
\centering\tablelayout
\caption{\textbf{Full-cohort component means.} Each entry is reference$\to$optimized, on the same NR scenes per family and interface. NC/DAC are multiplier penalties, EP/TTC/C enter the weighted average, and DDC is diagnostic with zero PDMS weight. PB: playback. TOAD uses Frozen$\to$CEM.}
\label{tab:interface-components}
\begin{tabular*}{\linewidth}{@{\extracolsep{\fill}}llrrrrrr@{}}
\toprule
Planner & Interface & NC & DAC & EP & TTC & C & DDC \\
\midrule
ReCogDrive-2B & PB & $0.92\!\to\!0.80$ & $0.86\!\to\!0.67$ & $0.59\!\to\!0.38$ & $0.90\!\to\!0.80$ & $1.00\!\to\!1.00$ & $0.97\!\to\!0.83$ \\
ReCogDrive-2B & LQR & $0.94\!\to\!0.94$ & $0.66\!\to\!0.91$ & $0.57\!\to\!0.76$ & $0.90\!\to\!0.91$ & $1.00\!\to\!1.00$ & $0.93\!\to\!0.99$ \\
ReCogDrive-8B & PB & $0.93\!\to\!0.84$ & $0.85\!\to\!0.70$ & $0.59\!\to\!0.40$ & $0.91\!\to\!0.85$ & $1.00\!\to\!1.00$ & $0.98\!\to\!0.85$ \\
ReCogDrive-8B & LQR & $0.95\!\to\!0.93$ & $0.70\!\to\!0.90$ & $0.63\!\to\!0.76$ & $0.91\!\to\!0.91$ & $1.00\!\to\!1.00$ & $0.95\!\to\!0.99$ \\
Qwen-Drive & PB & $0.96\!\to\!0.75$ & $0.91\!\to\!0.61$ & $0.68\!\to\!0.40$ & $0.96\!\to\!0.74$ & $1.00\!\to\!0.97$ & $0.99\!\to\!0.86$ \\
Qwen-Drive & LQR & $0.95\!\to\!0.98$ & $0.77\!\to\!0.91$ & $0.69\!\to\!0.76$ & $0.93\!\to\!0.98$ & $1.00\!\to\!1.00$ & $0.99\!\to\!0.99$ \\
NoRD & PB & $0.92\!\to\!0.93$ & $0.66\!\to\!0.80$ & $0.43\!\to\!0.61$ & $0.91\!\to\!0.91$ & $1.00\!\to\!1.00$ & $0.89\!\to\!0.97$ \\
NoRD & LQR & $0.92\!\to\!0.93$ & $0.55\!\to\!0.73$ & $0.44\!\to\!0.64$ & $0.89\!\to\!0.91$ & $1.00\!\to\!1.00$ & $0.84\!\to\!0.95$ \\
TOAD & PB & $0.99\!\to\!0.99$ & $0.86\!\to\!0.93$ & $0.69\!\to\!0.79$ & $0.94\!\to\!0.96$ & $1.00\!\to\!0.98$ & $0.99\!\to\!0.99$ \\
TOAD & LQR & $0.99\!\to\!0.99$ & $0.92\!\to\!0.99$ & $0.79\!\to\!0.86$ & $0.95\!\to\!0.97$ & $1.00\!\to\!1.00$ & $0.99\!\to\!1.00$ \\
\bottomrule
\end{tabular*}
\end{table}

\paragraph{Qwen speed revisions and alternative explanations.}
For each of the eight saved plans, vehicle-center positions are obtained from rear-axle poses with the pinned vehicle offset. Longitudinal speed uses the forward difference over 0.1\,s projected onto the current heading. Seven same-time revisions are averaged within each scene before aggregation. On the full scene set the means are $-0.273$ (IL) and $+2.167$ (RL) under playback, versus $-0.058$ and $+0.584$ under LQR. The paired LQR-minus-playback differences are $+0.215$ [$+0.180,+0.253$] for IL and $-1.584$ [$-1.710,-1.453$] for RL (log-bootstrap 95\% intervals). These are descriptive behavioral responses to the interface intervention, not a causal mediation estimate.

Two separately constructed Qwen controls help qualify the interpretation. At fixed original states and non-image inputs, replacing recorded images with renders changes PDMS by $-0.079$ (IL) and $-0.071$ (RL); the differential change is $+0.008$ [$-0.037,+0.050$], on the full paired scene set. This does not establish rendering equivalence at later visited states. Holding the same first predicted plan fixed, playback-minus-LQR changes are $+0.029$ [$-0.001,+0.060$] for IL and $-0.005$ [$-0.035,+0.024$] for RL; the differential change is $-0.034$ [$-0.075,+0.003$]. These contrasts are not added to or subtracted from the new replanning comparison as a causal decomposition. Together, they motivate examining execution and subsequent feedback jointly rather than attributing the rollout reversal solely to image replacement or single-plan smoothing.

\paragraph{Interpretation and remaining identification limits.}
The observed Goodhart-related concern is the scope of a score-based improvement claim: optimizing the tracked score need not improve requested-motion diagnostics or preserve gains under another execution interface. Tracking that makes a planner--controller system work well is not itself a benchmark failure. Nor does the observed interaction prove that training deliberately or causally selected a particular discarded component. Qwen-Drive's plan-adherence error improves after RL, and TOAD's increases slightly (Section~\ref{sec:interventions}); a universal increase in request--execution distance is not supported. NoRD's vocabulary and TOAD's penalties are possible explanations for their positive gains under both interfaces, not established causal protection mechanisms. Testing those explanations would require matched constraint ablations. The present intervention covers NR traffic and four-second failure-selected scenarios; its conclusions do not establish reactive or real-world transfer.

\subsection{Historical Playback Comparisons}
\label{app:closedloop-recog}
\label{app:closedloop-qwen}
\label{app:closedloop-nord}
\label{app:toad-closedloop}
\label{app:closedloop-failures}

\begin{figure}[!ht]
\centering
\includegraphics[width=\linewidth]{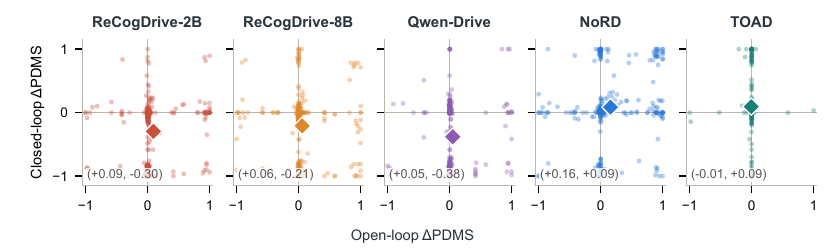}
\caption{\textbf{Historical playback gain transfer on the WorldEngine test set} \citep{worldengine}. Open-loop versus NR closed-loop PDMS changes (optimized minus baseline): IL$\to$RL for the first four panels, DrivoR$\to$TOAD for the last. Points: scenes; diamonds: paired means. Mean changes appear in Table~\ref{tab:closedloop-recog}.}
\label{fig:closedloop}
\end{figure}

Table~\ref{tab:closedloop-recog} preserves the earlier playback comparisons in both traffic regimes; Figure~\ref{fig:closedloop} shows their NR scene-level distributions. These are historical results, separate from the full matched execution intervention in Table~\ref{tab:execution-interface}.
Under this historical playback protocol, ReCogDrive and Qwen-Drive reverse their open-loop ordering, while NoRD retains a gain and TOAD improves despite a small negative open-loop change on the same scenes.

\newsavebox{\closedlooptablebox}
\begin{table}[!ht]
\centering
\captionsetup{skip=4pt}
\sbox{\closedlooptablebox}{\begin{minipage}{0.56\linewidth}
\centering\tablelayout
\setlength{\tabcolsep}{1.3pt}
\begin{tabular*}{\linewidth}{@{\extracolsep{\fill}}lrrr@{}}
\toprule
& \multicolumn{3}{c}{Paired $\Delta$PDMS $\uparrow$} \\
\cmidrule(lr){2-4}
Planner & OL & CL-NR & CL-R \\
\midrule
ReCogDrive-2B & \textcolor{tableimprove}{\textbf{+0.094}} & \textcolor{tabledegrade}{\textbf{\textminus{}0.296}} & \textcolor{tabledegrade}{\textbf{\textminus{}0.248}} \\
ReCogDrive-8B & \textcolor{tableimprove}{\textbf{+0.057}} & \textcolor{tabledegrade}{\textbf{\textminus{}0.207}} & \textcolor{tabledegrade}{\textbf{\textminus{}0.179}} \\
Qwen-Drive & \textcolor{tableimprove}{\textbf{+0.050}} & \textcolor{tabledegrade}{\textbf{\textminus{}0.380}} & \textcolor{tabledegrade}{\textbf{\textminus{}0.358}} \\
NoRD & \textcolor{tableimprove}{\textbf{+0.159}} & \textcolor{tableimprove}{\textbf{+0.087}} & \textcolor{tableimprove}{\textbf{+0.116}} \\
TOAD & \textcolor{tabledegrade}{\textbf{\textminus{}0.007}} & \textcolor{tableimprove}{\textbf{+0.094}} & \textcolor{tableimprove}{\textbf{+0.060}} \\
\bottomrule
\end{tabular*}
\end{minipage}}
\begin{minipage}[t]{0.56\linewidth}
\vspace{0pt}
\caption{\textbf{Historical playback gain transfer.} WorldEngine test set \citep{worldengine}; closed-loop execution uses playback. Open-loop (OL), non-reactive closed-loop (CL-NR), and reactive closed-loop (CL-R) changes. Regimes are matched separately.}
\label{tab:closedloop-recog}
\usebox{\closedlooptablebox}
\end{minipage}\hfill
\begin{minipage}[t]{0.415\linewidth}
\vspace{0pt}
\centering
\includegraphics[height=\dimexpr\ht\closedlooptablebox+\dp\closedlooptablebox\relax]{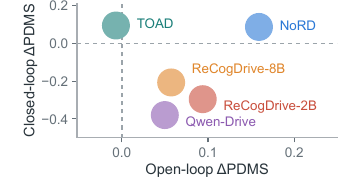}
\captionof{figure}{\textbf{Historical playback mean gains.} Each circle shows a planner's mean open-loop and non-reactive closed-loop score changes on matched scenes. Colors identify planners; circles have equal size.}
\label{fig:closedloop-means}
\end{minipage}
\end{table}

In NR evaluation, the counts of scenes with better/worse scores are 62/167 for ReCogDrive-2B, 67/164 for ReCogDrive-8B, 76/153 for Qwen-Drive, 134/83 for NoRD, and 138/51 for TOAD; ties are omitted.
Figure~\ref{fig:closedloop-means} uses the same NR-matched scenes and unnormalized PDMS changes. Each equal-size circle is centered on the observed paired mean; marker size does not encode uncertainty.

The NR ReCogDrive changes have paired-test $p$-values $2\times10^{-17}$ (2B) and $4\times10^{-12}$ (8B); paired contrasts between regimes have $p=4\times10^{-22}$ and $2\times10^{-12}$.
Qwen-Drive's paired scene-bootstrap intervals (seed 0) are $[-0.442,-0.318]$ (NR) and $[-0.417,-0.298]$ (R); its NR-matched open-loop interval is $[0.016,0.083]$.
NoRD's gains have $p=3\times10^{-4}$ (NR) and $10^{-6}$ (R).
TOAD's paired bootstrap intervals are $[0.053,0.139]$ (NR) and $[0.015,0.107]$ (R), with $p=5\times10^{-8}$ and $4\times10^{-5}$.

For ReCogDrive-2B, at-fault collisions rise from 15 to 63 scenes and drivable-area exits from 42 to 99; zero-score rollouts rise from 56 to 150.
ReCogDrive-8B's zero-score count rises from 73 to 128.
For Qwen-Drive, NR PDMS falls from $0.780$ to $0.400$: at-fault collisions rise from 11 to 72, drivable-area exits from 28 to 111, and zero-score rollouts from 39 to 165.
With reactive traffic, PDMS falls from $0.766$ to $0.408$, with collisions increasing from 13 to 57 and drivable-area exits from 28 to 118.
NoRD's zero-score count falls from 114 to 94, while TOAD's falls from 46 to 23 (NR) and 45 to 30 (R).
TOAD's NR drivable-area exits fall from 41 to 19, while at-fault collisions change from 5 to 4.

\paragraph{Implementation checks.}
\label{app:closedloop-verification}
All comparisons use the same adapter.
For the verified ReCogDrive pair, the first planning step reproduces the open-loop trajectory within 4\,cm for both checkpoints.
Re-executing 72 rollouts reproduces all seven score channels bit-identically.

\subsection{Successive Predictions and Rollout Diagnostics}
\label{app:execution-feedback}

\begin{table}[!ht]

\centering
\captionsetup{font=footnotesize,skip=4pt}
\caption{\textbf{ReCogDrive predicted foldbacks.} Historical WorldEngine playback with non-reactive traffic, 286 matched scenes; 0.05\,m filter. CL: closed loop. Rates over $n$ predictions (lower is better).}
\label{tab:feedback-foldbacks}
\tablelayout
\begin{tabular*}{\linewidth}{@{\extracolsep{\fill}}lrrr@{}}
\toprule
ReCogDrive & Open loop & First CL & All CL \\
& $n=286$ & $n=286$ & $n=2{,}288$ \\
\midrule
2B-IL & 0.0\% & 0.0\% & 17.9\% \\
2B-RL & 6.3\% & 6.3\% & 36.2\% \\
8B-IL & 0.3\% & 0.3\% & 7.6\% \\
8B-RL & 1.4\% & 0.7\% & 35.4\% \\
\bottomrule
\end{tabular*}

\end{table}

\paragraph{Qwen-Drive speed revisions.}
We analyze the complete, matched Qwen-Drive NR rollouts using the stored plans and states.
RL increases speed at the first update in 233 scenes and finishes more than 2\,m/s above its initial speed in 249; the corresponding IL counts are 101 and 26.
For NoRD, the corresponding RL counts are 94 and 33 on its matched set.

Equation~\ref{eq:speed-revision} in the main text decomposes the observed speed change into the previous plan's intended change and a same-time replanning revision. Speeds are extracted using the official playback conversion, with $h=0.5$\,s; the second term compares predictions at the same absolute future time.
Sign counts use a $10^{-8}$\,m/s numerical tolerance.
The maximum reconstruction error over the saved records is $3.7\times10^{-13}$\,m/s.
Across 2,009 adjacent-plan comparisons per checkpoint, the mean revision is $-0.262$\,m/s for IL and $+2.184$\,m/s for RL; 576 and 1,775 revisions are positive, respectively.
All seven revisions are positive in one IL scene and 196 RL scenes.
Meanwhile, 1,897/2,296 RL plans (82.6\%) predict a speed decrease from $h$ to $2h$, compared with 1,284 IL plans.
Using path length over the same future half-second interval instead of a local velocity difference retains the contrast: mean revisions are $-0.205$ and $+1.042$\,m/s, with all seven positive in 0 and 203 scenes.
These are descriptive within-scene sequences; their updates are not treated as independent experimental replicates.

\paragraph{Qwen-Drive case selection.}
Figure~\ref{fig:qwen-timeline} shows WorldEngine NR scene \texttt{02b68b9cc51f506a}.
We selected the first token in lexical order among 19 scenes with an open-loop gain above 0.01, IL closed-loop score above 0.8, RL closed-loop score zero, initial speed above 2\,m/s, and at least seven positive RL speed updates.
Across the eight updates, RL speed rises from 4.939 to 17.121\,m/s and the vehicle leaves the road at 4\,s; its closed-loop PDMS is zero, compared with 0.814 for IL.
The full matched-set counts above contextualize this illustrative case.

\paragraph{Continuation to the same junction.}
For the case in Figure~\ref{fig:qwen-timeline}, the saved IL continuation extends the rollout to 14\,s with 28 half-second decisions in total.
Replaying the original predictions reproduces the four-second state sequence exactly; 108 camera files across the two continuation-prefix checks are byte-identical.
The continuation uses the original Gaussian assets and later real nuPlan actor records, with fresh planner inference after the original horizon.
At 12\,s, IL has traveled 37.67\,m and is 3.59\,m from RL's four-second endpoint (38.32\,m traveled); its longitudinal speed is 3.25\,m/s versus RL's 17.12\,m/s.
All four IL footprint corners lie in the official nuPlan drivable area at all 29 saved half-second frames through 14\,s; only one RL corner remains inside at 4\,s.
These map checks describe road keeping at saved states; the original four-second scores remain unchanged, and no standard 14-second PDMS is assigned.
The continuation compares behavior near the same junction with different arrival times and speeds. Figure~\ref{fig:qwen-timeline} samples each row chronologically, using independent time grids to retain the continuation in the IL row.

\paragraph{ReCogDrive prediction foldbacks.}
\label{app:foldbacks}
We read native eight-point open-loop trajectories and the corresponding half-second knots of interpolated closed-loop plans.
Each point is relative to the current ego rear axle.
Define a prefix foldback by
\begin{equation}
I_\epsilon(u)=\mathbf1\!\left[x(h)>\epsilon\ \land\ x(h)-x(2h)>\epsilon\right].
\label{eq:foldback}
\end{equation}
We use $\epsilon=0.05$\,m to exclude small displacements and also report $0.5$\,m; these are descriptive amplitude filters, not safety boundaries.
Projecting the second displacement onto the predicted heading at $h$ retains every counted event at its respective threshold.
All four checkpoints have complete open-loop outputs on \texttt{navtest}.
For each NR matched set, we count the eight predictions used for state updates, excluding a terminal prediction that is saved but not executed.
Table~\ref{tab:foldback-counts} reports all configurations and both filters, including IL increases.
The matched-token open-loop and initial closed-loop records come from their respective runs; matching scene identity does not make their observations or predictions identical.

\begin{table}[!t]
\centering
\caption{\textbf{ReCogDrive foldbacks on \texttt{navtest} and the WorldEngine test set.} Each cell gives counts at $\epsilon=0.05$\,m / $0.5$\,m. Column denominators are per configuration. The last column counts scenes with at least one qualifying prediction; the preceding columns count predictions. No outcome-based filtering or new model inference is used.}
\label{tab:foldback-counts}
\tablelayout
\begin{tabular*}{\linewidth}{@{\extracolsep{\fill}}llrrrrr@{}}
\toprule
& & \texttt{navtest} & \multicolumn{4}{c}{WorldEngine test set (NR)} \\
\cmidrule(lr){3-3}\cmidrule(lr){4-7}
Planner & Variant & Full OL & Matched OL & First CL & All CL & Any CL \\
& Denominator & 12,146 & 286 & 286 & 2,288 & 286 \\
\midrule
ReCogDrive-2B & IL & 14 / 0 & 0 / 0 & 0 / 0 & 410 / 13 & 146 / 7 \\
 & RL & 522 / 31 & 18 / 2 & 18 / 1 & 829 / 528 & 226 / 197 \\
ReCogDrive-8B & IL & 13 / 0 & 1 / 0 & 1 / 0 & 174 / 1 & 68 / 1 \\
 & RL & 64 / 0 & 4 / 0 & 2 / 0 & 810 / 646 & 233 / 214 \\
\bottomrule
\end{tabular*}
\end{table}

\subsection{nuPlan Score Reproduction}
\label{app:nuplan-repro}

Plan-R1's reinforcement scores agree with the published CLS values within 0.1 points in both regimes (Table~\ref{tab:nuplan-repro}).
Recomputing scenario scores from stored sub-metrics agrees with the official aggregate to $10^{-15}$.

\begin{table}[H]
\caption{\textbf{nuPlan \texttt{val14}: score reproduction.} CLS in score points ($\times100$); a dash denotes no published value.}
\label{tab:nuplan-repro}
\centering
\tablelayout
\begin{tabular*}{\linewidth}{@{\extracolsep{\fill}}lrrrr@{}}
\toprule
& \multicolumn{2}{c}{Non-reactive} & \multicolumn{2}{c}{Reactive} \\
\cmidrule(lr){2-3}\cmidrule(lr){4-5}
Checkpoint & Published $\uparrow$ & Ours $\uparrow$ & Published $\uparrow$ & Ours $\uparrow$ \\
\midrule
PDM-Closed & $92.84$ & $92.84$ & $92.12$ & $92.13$ \\
Plan-R1 imitation & -- & $85.48$ & -- & $81.01$ \\
Plan-R1 reinforcement & $88.98$ & $88.89$ & $87.69$ & $87.71$ \\
\bottomrule
\end{tabular*}
\end{table}

\subsection{nuPlan Paired Scores and Motion}
\label{app:nuplan-ranking}

Tables~\ref{tab:nuplan-rank-nr} and~\ref{tab:nuplan-rank-r} retain all official score components and the motion diagnostics used in the main analysis.
Both traffic settings use the same \texttt{val14} scenarios; each retains the original correction family defined in Appendix~\ref{app:uncertainty}.
The score, progress, and plan-consistency improvements are reported alongside the increased absolute jerk peaks; unsupported changes remain identified rather than interpreted as effects.

\begin{table}[H]
\caption{\textbf{nuPlan \texttt{val14}: non-reactive traffic (NR).}
Plan-R1 imitation and reinforcement means, paired changes and uncertainty.
$^{\dagger}$ does not pass all three statistical checks of Appendix~\ref{app:uncertainty}.}
\label{tab:nuplan-rank-nr}
\centering
\tablelayout
\setlength{\tabcolsep}{2pt}
\begin{tabular*}{\linewidth}{@{\extracolsep{\fill}}lrrrrrr@{}}
\toprule
Quantity & IL & RL & $\Delta$ & Scene CI & Log CI & $p_{\rm Holm}$ \\
\midrule
CLS $\uparrow$ & $0.855$ & $0.889$ & $+0.034$ & \ci{0.018}{0.050} & \ci{0.019}{0.050} & $<10^{-4}$ \\
Drivable-area compliance $\uparrow$ & $0.942$ & $0.984$ & $+0.042$ & \ci{0.030}{0.054} & \ci{0.030}{0.056} & $<10^{-4}$ \\
No-at-fault-collision gate $\uparrow$ & $0.956$ & $0.953$ & $-0.004^{\dagger}$ & \ci{-0.016}{0.008} & \ci{-0.015}{0.008} & $1.000$ \\
Driving-direction gate $\uparrow$ & $0.989$ & $0.980$ & $-0.009^{\dagger}$ & \ci{-0.016}{-0.002} & \ci{-0.017}{-0.002} & $0.134$ \\
Making-progress gate $\uparrow$ & $0.986$ & $0.999$ & $+0.013$ & \ci{0.007}{0.021} & \ci{0.007}{0.020} & $0.002$ \\
Progress term $\uparrow$ & $0.908$ & $0.923$ & $+0.015$ & \ci{0.007}{0.023} & \ci{0.006}{0.023} & $0.022$ \\
TTC term $\uparrow$ & $0.901$ & $0.902$ & $+0.001^{\dagger}$ & \ci{-0.015}{0.018} & \ci{-0.017}{0.018} & $1.000$ \\
Speed-limit compliance $\uparrow$ & $0.980$ & $0.996$ & $+0.017$ & \ci{0.013}{0.021} & \ci{0.013}{0.020} & $<10^{-4}$ \\
Comfort term $\uparrow$ & $0.999$ & $0.997$ & $-0.002^{\dagger}$ & \ci{-0.005}{0.002} & \ci{-0.006}{0.002} & $1.000$ \\
Absolute lon.\ jerk peak [m/s$^3$] $\downarrow$ & $0.765$ & $0.901$ & $+0.136$ & \ci{0.110}{0.162} & \ci{0.107}{0.168} & $<10^{-4}$ \\
Absolute jerk peak [m/s$^3$] $\downarrow$ & $0.773$ & $0.841$ & $+0.068$ & \ci{0.043}{0.093} & \ci{0.043}{0.092} & $<10^{-4}$ \\
Planned heading jitter $\downarrow$ & $1.780$ & $1.546$ & $-0.235$ & \ci{-0.272}{-0.199} & \ci{-0.280}{-0.193} & $<10^{-4}$ \\
Smoothing displacement [m] $\downarrow$ & $0.068$ & $0.068$ & $-0.000^{\dagger}$ & \ci{-0.001}{0.001} & \ci{-0.001}{0.001} & $0.344$ \\
Plan churn [m] $\downarrow$ & $0.418$ & $0.382$ & $-0.037$ & \ci{-0.044}{-0.030} & \ci{-0.044}{-0.030} & $<10^{-4}$ \\
First-plan open-loop score $\uparrow$ & $0.753$ & $0.773$ & $+0.020$ & \ci{0.008}{0.032} & \ci{0.007}{0.032} & $<10^{-4}$ \\
\bottomrule
\end{tabular*}
\end{table}

\begin{table}[H]
\caption{\textbf{nuPlan \texttt{val14}: reactive traffic (R).}
Plan-R1 imitation and reinforcement means, paired changes and uncertainty.
$^{\dagger}$ does not pass all three statistical checks of Appendix~\ref{app:uncertainty}.}
\label{tab:nuplan-rank-r}
\centering
\tablelayout
\setlength{\tabcolsep}{2pt}
\begin{tabular*}{\linewidth}{@{\extracolsep{\fill}}lrrrrrr@{}}
\toprule
Quantity & IL & RL & $\Delta$ & Scene CI & Log CI & $p_{\rm Holm}$ \\
\midrule
CLS $\uparrow$ & $0.810$ & $0.877$ & $+0.067$ & \ci{0.051}{0.083} & \ci{0.052}{0.083} & $<10^{-4}$ \\
Drivable-area compliance $\uparrow$ & $0.948$ & $0.981$ & $+0.033$ & \ci{0.022}{0.045} & \ci{0.021}{0.047} & $<10^{-4}$ \\
No-at-fault-collision gate $\uparrow$ & $0.934$ & $0.944$ & $+0.010^{\dagger}$ & \ci{-0.003}{0.022} & \ci{-0.002}{0.021} & $0.598$ \\
Driving-direction gate $\uparrow$ & $0.988$ & $0.983$ & $-0.005^{\dagger}$ & \ci{-0.011}{0.001} & \ci{-0.012}{0.001} & $0.598$ \\
Making-progress gate $\uparrow$ & $0.960$ & $0.996$ & $+0.037$ & \ci{0.026}{0.048} & \ci{0.026}{0.047} & $<10^{-4}$ \\
Progress term $\uparrow$ & $0.851$ & $0.895$ & $+0.044$ & \ci{0.033}{0.055} & \ci{0.034}{0.055} & $<10^{-4}$ \\
TTC term $\uparrow$ & $0.888$ & $0.902$ & $+0.013^{\dagger}$ & \ci{-0.003}{0.029} & \ci{-0.004}{0.031} & $0.598$ \\
Speed-limit compliance $\uparrow$ & $0.982$ & $0.996$ & $+0.014$ & \ci{0.011}{0.019} & \ci{0.011}{0.018} & $<10^{-4}$ \\
Comfort term $\uparrow$ & $0.996$ & $0.987$ & $-0.009$ & \ci{-0.015}{-0.004} & \ci{-0.015}{-0.004} & $0.039$ \\
Absolute lon.\ jerk peak [m/s$^3$] $\downarrow$ & $0.845$ & $1.051$ & $+0.206$ & \ci{0.176}{0.235} & \ci{0.172}{0.242} & $<10^{-4}$ \\
Absolute jerk peak [m/s$^3$] $\downarrow$ & $0.812$ & $0.957$ & $+0.145$ & \ci{0.119}{0.171} & \ci{0.118}{0.173} & $<10^{-4}$ \\
Planned heading jitter $\downarrow$ & $1.437$ & $0.899$ & $-0.538$ & \ci{-0.610}{-0.471} & \ci{-0.612}{-0.463} & $<10^{-4}$ \\
Smoothing displacement [m] $\downarrow$ & $0.068$ & $0.071$ & $+0.003$ & \ci{0.002}{0.004} & \ci{0.002}{0.004} & $<10^{-4}$ \\
Plan churn [m] $\downarrow$ & $0.467$ & $0.448$ & $-0.019$ & \ci{-0.029}{-0.008} & \ci{-0.030}{-0.007} & $<10^{-4}$ \\
First-plan open-loop score $\uparrow$ & $0.745$ & $0.770$ & $+0.025$ & \ci{0.013}{0.036} & \ci{0.011}{0.037} & $<10^{-4}$ \\
\bottomrule
\end{tabular*}
\end{table}

\section{EPDMS Rescoring and Two-Stage Evaluation}
\label{app:v2-details}

\subsection{Fixed-Output EPDMS Rescoring}
\label{app:v2-replications}

Table~\ref{tab:v2pairs-uncertainty} collects the scene- and log-level intervals and adjusted tests for the fixed-output \texttt{navtest} changes in Table~\ref{tab:pairs}.
These single-stage readouts use v2 direction, lane keeping, and history comfort; v2 DDC excludes intersection positions and enters the score as a gate. The cross-frame and stage aggregation of the full two-stage experiment is evaluated separately below.

\begin{table}[!htbp]
\caption{
\textbf{EPDMS rescoring on \texttt{navtest}: paired statistical support.} For each pair, rows give Holm-adjusted Wilcoxon $p$-values, scene-bootstrap 95\% intervals, and driving-log-bootstrap 95\% intervals for the mean changes reported in Table~\ref{tab:pairs}. PDMS uses the v1 scorer; EPDMS, driving-direction compliance (DDC), lane keeping (LK), and history comfort (HC) use the v2 scorer on fixed outputs. Holm correction retains the original test families in Appendix~\ref{app:uncertainty}, including TOAD's augmentation-only arm, without recomputing tests for the displayed subset.
}
\label{tab:v2pairs-uncertainty}
\centering
\tablelayout
\begin{tabular*}{\linewidth}{@{\extracolsep{\fill}}lrrrrr@{}}
\toprule
& NAVSIM-v1 & \multicolumn{4}{c}{NAVSIM-v2 scorer} \\
\cmidrule(lr){2-2}\cmidrule(lr){3-6}
Pair & $\Delta$PDMS & $\Delta$EPDMS & $\Delta$DDC & $\Delta$LK & $\Delta$HC \\
\midrule
ReCog-2B IL$\to$RL
 & $<\!10^{-4}$ & $<\!10^{-4}$ & $<\!10^{-4}$ & $<\!10^{-4}$ & $<\!10^{-4}$ \\
{\scriptsize Scene CI} & \ci{.038}{.046} & \ci{.015}{.024} & \ci{-.012}{-.009} & \ci{-.063}{-.052} & \ci{-.007}{-.004} \\
{\scriptsize Log CI} & \ci{.035}{.050} & \ci{.009}{.029} & \ci{-.017}{-.006} & \ci{-.073}{-.043} & \ci{-.008}{-.004} \\
ReCog-8B IL$\to$RL
 & $<\!10^{-4}$ & $<\!10^{-4}$ & $<\!10^{-4}$ & $<\!10^{-4}$ & $<\!10^{-4}$ \\
{\scriptsize Scene CI} & \ci{.031}{.038} & \ci{.013}{.021} & \ci{-.006}{-.004} & \ci{-.029}{-.020} & \ci{-.005}{-.003} \\
{\scriptsize Log CI} & \ci{.029}{.041} & \ci{.010}{.024} & \ci{-.007}{-.003} & \ci{-.037}{-.014} & \ci{-.006}{-.002} \\
Qwen-Drive SFT$\to$RL
 & $<\!10^{-4}$ & $<\!10^{-4}$ & $0.00083$ & $<\!10^{-4}$ & $<\!10^{-4}$ \\
{\scriptsize Scene CI} & \ci{.023}{.029} & \ci{.016}{.021} & \ci{-.002}{-.000} & \ci{-.016}{-.008} & \ci{-.006}{-.003} \\
{\scriptsize Log CI} & \ci{.021}{.030} & \ci{.014}{.023} & \ci{-.002}{-.000} & \ci{-.019}{-.004} & \ci{-.007}{-.002} \\
DDv1$\to$v2
 & $<\!10^{-4}$ & $<\!10^{-4}$ & $<\!10^{-4}$ & $<\!10^{-4}$ & $<\!10^{-4}$ \\
{\scriptsize Scene CI} & \ci{.025}{.032} & \ci{-.001}{.007} & \ci{-.011}{-.008} & \ci{-.024}{-.016} & \ci{-.098}{-.087} \\
{\scriptsize Log CI} & \ci{.023}{.035} & \ci{-.005}{.011} & \ci{-.015}{-.005} & \ci{-.028}{-.012} & \ci{-.107}{-.079} \\
\midrule
NoRD IL$\to$RL
 & $<\!10^{-4}$ & $<\!10^{-4}$ & $<\!10^{-4}$ & $<\!10^{-4}$ & $<\!10^{-4}$ \\
{\scriptsize Scene CI} & \ci{.101}{.114} & \ci{.091}{.104} & \ci{.012}{.017} & \ci{.008}{.019} & \ci{.017}{.024} \\
{\scriptsize Log CI} & \ci{.096}{.120} & \ci{.086}{.109} & \ci{.010}{.019} & \ci{.005}{.023} & \ci{.017}{.025} \\
\midrule
GTRS-Dense expert$\to$reward
 & $<\!10^{-4}$ & $<\!10^{-4}$ & $0.034$ & $<\!10^{-4}$ & $<\!10^{-4}$ \\
{\scriptsize Scene CI} & \ci{.007}{.014} & \ci{-.001}{.007} & \ci{-.002}{-.000} & \ci{-.013}{-.005} & \ci{.006}{.010} \\
{\scriptsize Log CI} & \ci{.005}{.016} & \ci{-.002}{.008} & \ci{-.002}{.000} & \ci{-.014}{-.004} & \ci{.004}{.013} \\
\midrule
DrivoR$\to$TOAD & $0.79$ & $<\!10^{-4}$ & $<\!10^{-4}$ & $<\!10^{-4}$ & $0.42$ \\
{\scriptsize Scene CI} & \ci{.000}{.002} & \ci{-.004}{-.002} & \ci{-.004}{-.002} & \ci{-.010}{-.006} & \ci{-.002}{.000} \\
{\scriptsize Log CI} & \ci{-.000}{.002} & \ci{-.005}{-.001} & \ci{-.005}{-.001} & \ci{-.012}{-.005} & \ci{-.002}{.000} \\
\bottomrule
\end{tabular*}
\end{table}

Table~\ref{tab:pairs-replicated} in Appendix~\ref{app:paired-replications} reports EPDMS rescoring on \texttt{navhard} and the WorldEngine test set alongside the original PDMS and diagnostic changes.
NoRD improves the revised score and displayed components on both sets.
ReCogDrive retains score gains, while some component changes have weaker support on the smaller sets.
For Qwen-Drive, the \texttt{navtest} lane-keeping and history-comfort declines are not significant on either smaller set.
These are evaluations of fixed outputs, separate from optimizing the revised objective.

\subsection{\rev{Reinforcement Pairs under the Two-Stage Protocol}}
\label{app:pairs-twostage}

\rev{We run the official NAVSIM-v2 \texttt{navhard-two-stage} protocol on 450 original scenes and 5{,}462 synthetic scenes for the ReCogDrive, Qwen-Drive, and NoRD pairs.
The protocol is a re-inference, not a rescoring: the 5{,}462 synthetic scenes start from counterfactual ego states and are rendered from those states, so each planner produces new trajectories from the rendered eight-camera observations (the synthetic assets contain no LiDAR, which excludes DiffusionDrive and DiffusionDriveV2, whose feature extractors consume the point cloud).
The planners run outside the v2 evaluation code through the same adapters as the closed-loop experiments: Qwen-Drive receives the three camera views at the four most recent frames and a history interpolated from the four ego states, ReCogDrive receives the front frame and the four ego statuses, and NoRD runs its released agent against its served model; on fourteen synthetic scenes whose navigation command is undefined, ReCogDrive is given the straight-ahead command that NoRD's own agent substitutes in the same case.
Predictions are then scored and aggregated by the unmodified official code, with the mapping between original and synthetic scenes taken over the full split.
Table~\ref{tab:twostage-main} reports scene-mean components for each stage and the official combined EPDMS. Paired bootstrap 95\% intervals resample the 450 original scenes for stage one and the 450 official mapping groups for stage two and the combined score. The stage-two score changes below use group-weighted means, which can differ from unweighted scene means.}

\rev{The three free-form pairs gain $0.105$, $0.079$, and $0.063$ PDMS on the same 450 scenes in open loop and $0.077$, $0.060$, and $0.062$ EPDMS under fixed-output rescoring (Table~\ref{tab:pairs-replicated}).
Under the two-stage protocol their stage-one changes are $+0.000$, $-0.015$, and $-0.003$: the original scenes are re-planned from rendered rather than recorded frames, and the open-loop advantage does not survive the re-inference.
On the synthetic stage the changes are $-0.012$, $-0.016$, and $-0.010$, with intervals containing zero. The combined changes are $-0.008$ $[-0.038,+0.019]$, $-0.008$ $[-0.035,+0.018]$, and $+0.003$ $[-0.021,+0.027]$, respectively; these intervals include zero and are marked $^{\dagger}$ in Table~\ref{tab:twostage-main}.
NoRD gains $+0.143$ and $+0.049$ in stages one and two, and $+0.079$ $[+0.048,+0.112]$ in combined EPDMS.}

\rev{Table~\ref{tab:pairs-twostage-components} decomposes the synthetic stage.
For both ReCogDrive pairs, DAC and EP increase while DDC decreases; Qwen-Drive improves all displayed components except EC.
The term that moves most is two-frame extended comfort. Within each stage, it compares time-aligned simulated motion from consecutive paired frames. On the synthetic stage it falls by $0.378$, $0.527$, and $0.296$, from $0.65$--$0.69$ to $0.15$--$0.35$, while NoRD's rises by $0.038$.
The reinforcement checkpoints therefore pass the cross-frame motion-consistency checks less often, complementing the closed-loop prediction-revision measurements in Appendix~\ref{app:execution-feedback}.}

\rev{For original-scene scores $S_{1,g}$ and weighted synthetic-child scores $\bar S_{2,g}$, the official combined score is $\mathbb{E}_g[S_{1,g}\bar S_{2,g}]$ rather than $\mathbb{E}_g[S_{1,g}]\mathbb{E}_g[\bar S_{2,g}]$, so changes in the two stage means need not share the direction of the combined change.}

\begin{table}[H]
\caption{
\textbf{NAVSIM-v2 \texttt{navhard-two-stage}: component analysis.} Synthetic-stage paired changes of the EPDMS components. EC is the two-frame extended comfort.
}
\label{tab:pairs-twostage-components}
\centering
\tablelayout
\begin{tabularx}{\linewidth}{@{}Xrrrrrrrr@{}}
\toprule
& \multicolumn{8}{c}{Stage 2 component changes $\uparrow$} \\
\cmidrule(lr){2-9}
Pair & $\Delta$NC & $\Delta$DAC & $\Delta$DDC & $\Delta$EP & $\Delta$TTC & $\Delta$LK & $\Delta$HC & $\Delta$EC \\
\midrule
ReCog-2B IL$\to$RL & $-.001$ & $+.005$ & $-.011$ & $+.061$ & $-.001$ & $+.006$ & $-.000$ & $-.378$ \\
ReCog-8B IL$\to$RL & $-.000$ & $+.008$ & $-.002$ & $+.056$ & $-.003$ & $+.003$ & $-.002$ & $-.527$ \\
Qwen-Drive SFT$\to$RL & $+.005$ & $+.029$ & $+.008$ & $+.006$ & $+.005$ & $+.005$ & $+.008$ & $-.296$ \\
NoRD SFT$\to$RL (vocab.) & $+.024$ & $+.120$ & $+.051$ & $+.040$ & $+.022$ & $+.024$ & $+.017$ & $+.038$ \\
\bottomrule
\end{tabularx}
\end{table}

\end{document}